\documentclass[10pt,twocolumn,letterpaper]{article}

\usepackage[pagenumbers]{cvpr}        %

\usepackage{cuted}    %
\usepackage{currfile} %
\usepackage{caption}  %
\usepackage{multirow}
\usepackage{pifont}
\usepackage{amsthm}
\usepackage{algorithm}
\usepackage{algpseudocode}
\newtheorem{theorem}{Theorem}
\newtheorem{proposition}{Proposition}

\theoremstyle{definition}

\theoremstyle{plain}

\makeatletter
\def\LN@posrec#1#2{\expandafter\gdef\csname LN@x@#1\endcsname{#2}}
\AtEndDocument{\let\LN@posrec\@gobbletwo}
\newcount\LN@seq
\newcommand*\makeLineNumberOuterColumn{%
  \global\advance\LN@seq\@ne
  \pdfsavepos
  \edef\LN@tmp{\write\@auxout{\noexpand\string\noexpand\LN@posrec
    {\the\LN@seq}{\noexpand\the\noexpand\pdflastxpos}}}\LN@tmp
  \expandafter\ifx\csname LN@x@\the\LN@seq\endcsname\relax
    \if@firstcolumn\makeLineNumberLeft\else\makeLineNumberRight\fi
  \else
    \ifdim\csname LN@x@\the\LN@seq\endcsname sp<.5\paperwidth
      \makeLineNumberLeft\else\makeLineNumberRight\fi
  \fi}
\AtBeginDocument{%
  
  }
\makeatother

\definecolor{cvprblue}{rgb}{0.21,0.49,0.74}
\usepackage[breaklinks,colorlinks,allcolors=cvprblue]{hyperref}
\newcommand{\Pz}{P_0}
\newcommand{\Pt}{P_\theta}
\newcommand{\st}{\sigma_\theta}
\newcommand{\sq}{\sigma_q}
\newcommand{\Gw}{G_\omega}
\newcommand{\Dp}{D_\psi}
\newcommand{\Ep}{E_\psi}
\newcommand{\qp}{q_\phi}
\newcommand{\zreach}{z^{\mathrm{reach}}}
\newcommand{\zhat}{\hat z}
\newcommand{\zgoal}{\hat z_g}
\newcommand{\pik}[1]{\pi_{#1}}
\newcommand{\rfifty}{r_{50}}
\newcommand{\replay}{\mathcal{B}}
\newcommand{\Zc}{\mathcal{Z}}
\newcommand{\Lam}{\Lambda}
\newcommand{\Pstar}{P^\star}
\newcommand{\Ball}{\mathbb{B}_\varepsilon}
\newcommand{\Fc}{\mathcal{F}}
\newcommand{\Ex}{\mathbb{E}}
\DeclareMathOperator{\med}{med}

\newcommand{\dimE}{\dim_{E}}
\newcommand{\cmark}{\ding{51}}
\newcommand{\xmark}{\textcolor{black!40}{\ding{55}}}

\hypersetup{pdftitle={Recursive Self-Models for Efficient and Long-Horizon JEPA Planning},pdfauthor={Zhewen He, Junyi Hu, Jiazhao Liang, Haomian Huang, Zhenhua Li, Shuaihang Yuan, Yi Fang}}

\def\confName{CVPR}
\def\confYear{2027}

\title{Recursive Self-Models for Efficient and Long-Horizon JEPA Planning}

\author{
Zhewen He \quad Junyi Hu \quad Jiazhao Liang \quad Haomian Huang \quad Zhenhua Li \quad Shuaihang Yuan \quad Yi Fang\thanks{Corresponding author.}\\
New York University
}

\begin{document}
\maketitle
\begin{strip}
  \centering
  \includegraphics[width=\textwidth]{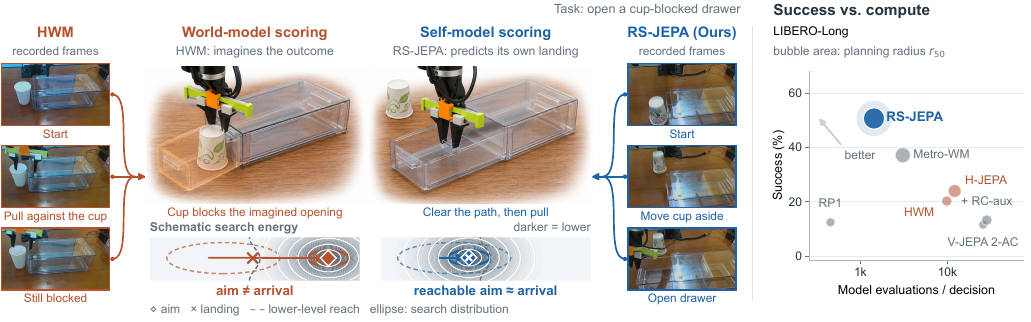}
  \captionsetup{hypcap=false}
  \captionof{figure}{\textbf{Aim versus arrival.} HWM pulls a cup-blocked drawer toward an unreachable aim; RS-JEPA first moves the cup aside, then opens the drawer. Frames from real robot recordings are shown on either side of the scene illustrations. Below, energy fields illustrate the aim--arrive gap and reachable subgoals. In the LIBERO-Long plot, bubble area denotes planning radius.}
  \captionsetup{hypcap=true}
  \label{fig:teaser}
\end{strip}

\begin{abstract}
Latent world models plan by searching for actions whose imagined outcome approaches a goal, but this signal is reliable only a few steps ahead. Hierarchical planners extend the reach by proposing subgoals with a coarser world model, yet the lower-level planner often cannot reach those subgoals. We argue that the upper level should model where \emph{its own lower-level planner} lands when pursuing a subgoal, rather than how the world evolves under abstract actions. We call this a self-model, and the resulting recursive planner RS-JEPA. Since reach can only be measured by execution, the self-improvement loop trains the self-model only on executed pursuits and their hindsight relabels, and the hierarchy improves with use. In an idealized setting with deterministic dynamics, a perfect world model and exact search, we prove that an optimistic self-model collapses a hierarchy of any depth to its flat planner and that a calibrated hierarchy of depth $K$ has the reach of a flat planner whose horizon is exponential in $K$ at compute linear in $K$. Training on executed outcomes provably reaches a unique fixed point within $K$ rounds, while training on imagined outcomes leaves an optimistic self-model in place. Because each level searches a short subproblem the level below can solve, and a prior learned from past plans warm-starts that search, the planner needs far fewer model evaluations per decision than flat search. The method applies to any JEPA world model, and we instantiate it on V-JEPA 2-AC and LeWM. Under the language-conditioned protocol for LIBERO and RoboCasa, with about 10{,}000 self-improvement episodes budgeted per suite, RS-JEPA reaches 50.7\% success on LIBERO-Long, with about 44$\times$ fewer model evaluations per decision than V-JEPA 2-AC at equal success and 18$\times$ fewer at their operating points. Within the same system, a world-model upper level in place of the self-model loses most of the gain. On a real robot, self-improvement also raises success.
\end{abstract}

\section{Introduction}
\label{sec:intro}

Latent world models make planning from pixels practical: encode the observation and the goal, then search for actions whose imagined outcome approaches the goal in latent space~\cite{assran2025vjepa2,zhou2025dinowm,sobal2025pldm,maes2026lewm}. This signal is reliable only a few steps ahead. On Meta-World, a predictor trained on 5-step rollouts ranks actions reliably only for nearby targets, while task goals lie up to 53 steps away, and the limit persists with a perfect simulator~\cite{alrasheed2026planning}. On two LIBERO scenes, the original LeWM planner completes 0\% of full episodes~\cite{cheng2026sage}.

Hierarchy is the usual remedy, and recent planners stack a coarser world model above the fine one~\cite{zhang2026hwm,zhang2026hjepa}. The upper level proposes subgoals by predicting how the world evolves under abstract actions. Yet the lower-level planner often cannot reach those subgoals~\cite{caselli2026hilewm,lee2026metro}. On PushT at offset 50, naive Hi-LeWM reaches 38.7\%, below the 52.7\% of flat LeWM, while true future latents as subgoals reach 73.3\%~\cite{caselli2026hilewm}. The lower level is therefore capable, and the loss comes from subgoal proposal. Upper-level search selects macro-actions that score well under the model but make poor control targets.

We argue that the upper level should model a different object: where \emph{its own lower-level planner} ends up when pursuing a subgoal. We call this a \emph{self-model} (\cref{fig:teaser}). Scoring a plan by predicted landing points makes the upper level choose stepping stones the planner can reach. What a planner can reach is revealed by executing it, so, as in hindsight hierarchical RL~\cite{levy2019hac}, the self-improvement loop trains the self-model only on executed pursuits and their hindsight relabels, and the hierarchy improves as it is used. Each level searches a short subproblem the level below can solve, so the planner also reaches farther with less compute. Applying the self-model recursively across levels gives RS-JEPA.

Our contributions are:
\begin{itemize}
  \item A \textbf{self-model} upper level: an option model~\cite{sutton1999options} of the search-based planner below that predicts its executed landing and, shared across levels, scores CEM plans over low-dimensional subgoal codes on any JEPA world model (\cref{sec:selfmodel,sec:planning}).
  \item \textbf{Grounded self-improvement}: a loop that trains the self-model only on executed outcomes and extends the planning radius over training (\cref{sec:loop}).
  \item A \textbf{planning-radius analysis} in an idealized setting. The latent metric limits flat planning even with a perfect model, and optimism collapses the hierarchy. A calibrated hierarchy multiplies the radius by $M$ per level at compute linear in depth. Subgoal search has a critical dimension, and executed training reaches a unique fixed point while imagined training stays optimistic (\cref{sec:analysis}).
\end{itemize}
On LIBERO~\cite{liu2023libero} and RoboCasa~\cite{nasiriany2024robocasa} under the standard language-conditioned evaluation protocol, RS-JEPA leads on four of five suites, with the largest gain on LIBERO-Long (\cref{sec:main}). Replacing the self-model with a world-model upper level loses most of this gain (\cref{sec:ablation}).

\section{Related Work}
\label{sec:related}

\paragraph{Latent World-Model Planning.}
Learned world models support policy learning~\cite{hafner2025dreamer}, latent trajectory optimization~\cite{hansen2024tdmpc2} and navigation planning in video space~\cite{bar2025nwm}. V-JEPA 2-AC~\cite{assran2025vjepa2}, DINO-WM~\cite{zhou2025dinowm}, PLDM~\cite{sobal2025pldm} and LeWM~\cite{maes2026lewm} plan by sampling-based search over actions to minimize latent distance to a goal, and Terver et al.~\cite{terver2026jepawm} study which design choices drive their success. Recent analyses locate the long-horizon limit outside predictor accuracy: it persists under perfect prediction~\cite{alrasheed2026planning}, stems from the planning objective~\cite{singh2026objective}, and defeats finite-budget search even with a perfect world model~\cite{gao2026imwm}. Reachability-aware objectives~\cite{singh2026objective,li2026rcaux} reshape the flat energy toward control distance, and such an energy can serve as Level 0 of RS-JEPA, whose upper levels then extend its reach.

\paragraph{Hierarchical Latent Planning.}
HWM~\cite{zhang2026hwm} learns world models at several temporal scales in a shared latent space, taking real Franka pick-and-place from 0\% to 70\% and reducing planning compute up to threefold in simulation. H-JEPA~\cite{zhang2026hjepa} trains a three-level hierarchy of JEPAs end to end, each in its own latent space. Hi-LeWM~\cite{caselli2026hilewm} identifies subgoal generation as the bottleneck and constrains search around macro-actions from training trajectories. ID-WM~\cite{zhou2026idwm} samples reachable transitions at the upper level, decoupled from local control. FF-JEPA~\cite{masip2026ffjepa}, LAGO~\cite{barbeau2026lago}, SAGE~\cite{cheng2026sage} and Xiang and Liu~\cite{xiang2026frozen} generate subgoals with networks trained on offline trajectories; PILOT~\cite{liu2026pilot} adds policy-initialized CEM. Metro-WM~\cite{lee2026metro} measures how often such subgoals are unrealizable and retrieves real frames from a graph over prior experience instead, and SURVEYOR~\cite{ahmad2026surveyor} skips subgoals when a calibrated tolerance deems the goal reachable. In all of these, the upper level is trained offline to model the world or the data, whereas that of RS-JEPA models the planner beneath it and trains on what that planner does.

\paragraph{Amortized Planners.}
GC-IDM~\cite{nguyen2026gcidm}, LeFlow~\cite{huang2026leflow}, LePlanner~\cite{bansal2026leplanner} and RP1~\cite{sommer2026rp1} replace, warm-start or amortize search with learned networks. They reduce the cost of a planning problem at a fixed horizon and can serve as Level 0 of a hierarchy, whereas we extend the radius.

\paragraph{Outcome and Reachability Models.}
An option model predicts the state in which a temporally extended action terminates~\cite{sutton1999options,yao2014uom}, and a self-model is an option model whose option is a planner pursuing a subgoal. SkiMo~\cite{shi2022skimo}, DADS~\cite{sharma2020dads} and LMTA~\cite{wang2026lmta} predict the outcome of learned skills or subgoal-conditioned workers, LEAP~\cite{nasiriany2019leap} judges subgoal reachability with a goal-conditioned value, and RC-aux~\cite{li2026rcaux} adds reachability supervision to a latent world model. Hierarchical reinforcement learning grounds the upper level in what the lower level achieves: HAC~\cite{levy2019hac} relabels each subgoal with the state actually reached and penalizes subgoals the current lower level misses, HRAC~\cite{zhang2020hrac} restricts subgoals to a learned $k$-step adjacency region, and SSE~\cite{hwang2026sse} separates unreachable from admissible subgoals using failed transitions; Director~\cite{hafner2022director} and HIQL~\cite{park2023hiql} learn subgoal and worker policies using learned value functions. Recursive planners split a trajectory at predicted midpoints and reach horizons exponential in depth~\cite{jurgenson2019subgoal,parascandolo2020dcmcts,pertsch2020gcp}, and subgoal search keeps the generated subgoals that a low-level policy reaches~\cite{czechowski2021ksubs}. RS-JEPA shares the executed-outcome grounding of HAC and SSE and the recursion of these planners. It differs in what the upper level learns: a predictor of where a search-based planner over a frozen world model lands, shared across levels and scored inside CEM at every level, rather than a value function or a policy over a learned worker. Sec.~D of the supplementary material compares the closest methods. In robotics, the term self-model usually refers to a model of the robot's body~\cite{bongard2006resilient,chen2022selfmodel}, and we use it for a model of the robot's planning ability.

\paragraph{Self-Improvement and Exploitation.}
DayDreamer~\cite{wu2023daydreamer} and SimDist~\cite{levy2026simdist} adapt world models on real robots, RISE~\cite{yang2026rise} improves a policy on imagined rollouts, World Action Verifier~\cite{liu2026wav} verifies subgoals through cycle consistency, and RSI-Anything~\cite{cui2026rsi} retains policy and world-model updates across domains. MBPO~\cite{janner2019mbpo} studies how far to trust model-generated data, and Bhamidipaty et al.~\cite{bhamidipaty2026exploitable} prove that imperfect world models are exploitable on large policy sets; at the upper level, executed outcomes correct this bias.

\section{Method}
\label{sec:method}

\begin{figure*}[t]
  \centering
  \includegraphics[width=\textwidth]{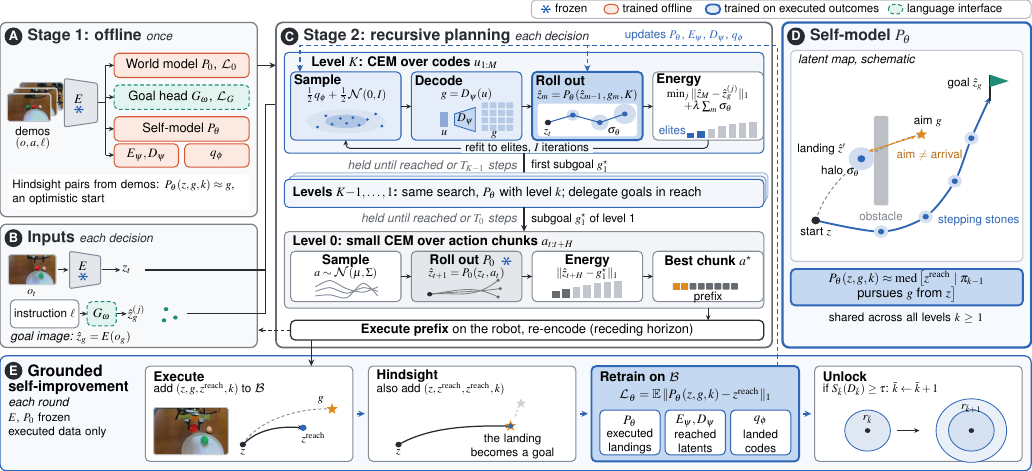}
  \caption{\textbf{RS-JEPA pipeline.} (A) Stage 1 trains the world model and pretrains the other modules on demonstrations. (B) Each decision encodes the observation and the goal. (C) Upper levels search subgoal codes scored by the self-model and hold the first subgoal until the level below reaches it or exhausts its budget; Level 0 searches action chunks with the frozen world model. (D) The self-model predicts where the lower-level planner lands. (E) Each round retrains on executed outcomes only.}
  \label{fig:pipeline}
\end{figure*}

RS-JEPA adds a self-model and a self-improvement loop to a frozen JEPA world model. \Cref{fig:pipeline} shows the pipeline, and supplementary Sec.~B lists the loop as an algorithm.

\subsection{Setting and Backbone}
\label{sec:setting}

Given an observation $o_t$ and a goal, as a language instruction $\ell$ or a goal image $o_g$, the agent outputs action chunks in closed loop, planning entirely in the latent space of a visual encoder $E$, with $z_t = E(o_t)$.

Level 0 is the action-conditioned predictor of the JEPA world model,
\begin{equation}
  \zhat_{t+1} = \Pz(z_t, a_t).
  \label{eq:wm}
\end{equation}
Because the method calls only $E$ and $\Pz$, any latent world model can serve as the backbone. Our main backbone is V-JEPA 2-AC~\cite{assran2025vjepa2} with its pretrained $E$, and $\Pz$ is trained on the benchmark demonstrations, since the published predictor is trained on DROID~\cite{khazatsky2024droid}, whose cameras, scenes and action distribution differ from the simulated benchmarks. \Cref{eq:wm} is trained with teacher-forcing and rollout losses over $L$ steps against stop-gradient target latents,
\begin{equation}
  \mathcal{L}_0 = \sum_t \big\| \Pz(z_t, a_t) - z_{t+1} \big\|_1 + \sum_t \big\| \zhat_{t+L} - z_{t+L} \big\|_1 ,
  \label{eq:wmloss}
\end{equation}
with $\zhat_{t+L}$ rolled out from $z_t$ under $a_{t:t+L-1}$. On LeWM~\cite{maes2026lewm}, $E$ and $\Pz$ are trained jointly from pixels instead of with \cref{eq:wmloss}.

Stage 1 runs once on demonstrations. Stage 2 is deployment and self-improvement, and it keeps $E$ and $\Pz$ frozen, so every improvement in the loop comes from the upper levels.

\subsection{Self-Model}
\label{sec:selfmodel}

A level-$k$ planner $\pik{k}$ receives a goal, plans, and hands subgoals to $\pik{k-1}$. The self-model predicts where the level below ends up:
\begin{equation}
  \begin{aligned}
    \zhat' &= \Pt(z, g, k) \\
    &\approx \med\big[\zreach \mid \pik{k-1} \text{ pursues } g \text{ from } z\big].
  \end{aligned}
  \label{eq:selfmodel}
\end{equation}
In \cref{eq:selfmodel}, $\zreach$ is the latent observed when $\pik{k-1}$ arrives at $g$ or exhausts its budget of $T_{k-1}$ steps, and $\med$ is the coordinate-wise conditional median, the minimizer of the $\ell_1$ training loss of \cref{sec:loop}. The same predictor serves all levels $k \ge 1$ through a level embedding, so the parameter count does not grow with depth. It is initialized from the weights of $\Pz$, with a subgoal embedding in place of the action embedding, and has several output heads trained on bootstrapped tuples, whose disagreement gives the uncertainty $\st(z, g, k)$~\cite{lakshminarayanan2017ensembles}.

The difference from a world-model upper level is the prediction target. A macro-action predictor maps a state and an abstract action to the next state, and its input can encode motions the lower level would never produce. The self-model predicts what this planner, with its current ability, achieves from $z$ when asked for $g$, a target that moves as the planner below improves. A subgoal behind an obstacle that $\pik{k-1}$ cannot round yields a predicted landing in front of the obstacle, so the energy of the plan reflects where the agent will actually be.

Weight sharing lets what lower levels learn about the scene carry upward, and the level embedding captures how reach differs between levels. A newly unlocked level starts collapsed onto the level below (\cref{sec:analysis}) and gains reach from its first round of data. The budget $T_k$ grows with the level, since a level-$k$ pursuit spans several subgoals of the level below.

\paragraph{Subgoal Codes.}
Subgoals are low-dimensional codes $u \in \mathbb{R}^{d_u}$ decoded by $g = \Dp(u)$, with $d_u$ between 4 and 16, rather than encoder tokens. An autoencoder $(\Ep, \Dp)$ on encoder latents provides the code space, and codes are standardized so that the unit Gaussian $\mathcal{N}(0, I)$ matches their marginal. The dimension $M d_u$ of a plan of $M$ codes decides whether search receives any signal (\cref{eq:critical}).

\subsection{Recursive Planning}
\label{sec:planning}

Level $K$ searches $M$ subgoal codes with the cross-entropy method (CEM)~\cite{rubinstein1999cem,chua2018pets}, rolling the self-model forward and scoring the predicted landing:
\begin{equation}
\begin{aligned}
  \min_{u_{1:M}}\; & \min_j \big\| \zhat_M - \zgoal^{(j)} \big\|_1 + \lambda \sum_{m=1}^{M} \st\big(\zhat_{m-1}, g_m, K\big), \\
  \text{s.t.}\;\; & \zhat_m = \Pt\big(\zhat_{m-1}, g_m, K\big), \;\; g_m = \Dp(u_m), \;\; \zhat_0 = z_t.
\end{aligned}
\label{eq:energy}
\end{equation}
CEM samples $N$ candidates and refits its sampling distribution to the elites for $I$ iterations. Unlike the standard CEM of V-JEPA 2-AC, half of the samples come from a learned prior $\qp(u \mid z, \zgoal, k)$ and half from the unit Gaussian over codes. The energy scores predicted landings, and its uncertainty term steers search away from subgoals the planner has never reached. Searching subgoal codes rather than raw actions lets each added level multiply the reach.

The first subgoal $g_1^\star$ of the elite plan becomes the goal of level $K{-}1$, which solves the same problem with its own level embedding. A level delegates the goal itself when its self-model predicts no better landing through intermediate subgoals. Level 0 searches an action chunk with $\Pz$, minimizing $\|\zhat_{t+H} - g_1^\star\|_1$. The agent executes a prefix of the chunk, re-encodes, and replans Level 0. An upper level holds its subgoal until the level below arrives or exhausts its budget, and then replans from the new observation.

This structure is the source of the efficiency (\cref{sec:radius}). A flat planner aiming at a distant goal spends most of its samples where the energy carries no signal. In RS-JEPA, the goal at every level lies within the radius of the level below, so Level 0 needs only a small CEM, an upper level ranks a few low-dimensional codes around a learned prior, and upper levels run rarely.

\subsection{Grounded Self-Improvement}
\label{sec:loop}

The self-improvement loop trains the self-model only on executed outcomes.

\paragraph{Initialization.}
Stage 1 pretrains $\Pt$ on hindsight pairs from demonstrations, with latents $z_t$ and $z_{t+h}$ as start and subgoal and $z_{t+h}$ as landing. Demonstrations always reach their goals, so the pretrained self-model is optimistic, $\Pt(z, g, k) \approx g$, and the initial recursive planner behaves like its flat planner (\cref{sec:analysis}). The prior $\qp$ is pretrained on the codes of demonstration subgoals.

\paragraph{Rounds.}
Each round runs four steps. \emph{Execute} attempts tasks with the current planner and records $(z, g, \zreach, k)$ at every level, and with a small probability an upper level takes its top plan in place of delegating. \emph{Hindsight} relabels each landing as a goal the planner has reached~\cite{andrychowicz2017her,ghosh2021gcsl}. The self-model thus predicts arrival where the planner has been and a shortfall where it has fallen short, which departs from the optimistic initialization and gives search its signal. Where a hindsight label overstates arrival, the executed tuple recorded once search aims there corrects it. \emph{Retrain} minimizes
\begin{equation}
  \mathcal{L}_\theta = \mathbb{E}_{(z, g, \zreach, k) \sim \replay} \big\| \Pt(z, g, k) - \zreach \big\|_1 ,
  \label{eq:loss}
\end{equation}
fine-tunes $(\Ep, \Dp)$ on demonstration and reached latents, which keeps decoded subgoals on the manifold of states the agent has visited, and fits $\qp$ to hindsight codes and to first subgoals whose executed landing matched the aim, narrowing its width $\sq$. \emph{Unlock} opens level $k{+}1$ once level $k$ succeeds at a target offset $D_k$ with rate $\tau$, giving a curriculum driven by ability, not a fixed schedule. Planning starts at the highest unlocked level $\bar k \le K$, written $K$ in \cref{eq:energy}.

Lower levels stay fixed during a round, which keeps its labels stationary. Across rounds, level-1 labels stay stationary because $\Pz$ is frozen, while labels for $k \ge 2$ shift as $\pik{k-1}$ improves. Batches are balanced over levels, and older tuples, which understate the current lower level, are down-weighted. Labels of a level depend only on the levels below, so the idealized round update is triangular, with a unique fixed point (\cref{thm:fixed}).

\paragraph{Correction of Selection Bias.}
Search concentrates on inputs whose energy the model underestimates, the optimizer's curse~\cite{smith2006curse}. Imagined landings carry this bias into the labels, whereas executed landings contradict it at exactly the subgoals whose predicted landing was too optimistic. Search thus yields informative data (\cref{sec:analysis}).

\paragraph{Evaluation Quantities.}
Besides task success, the planning radius $\rfifty$ is the farthest goal offset, in control steps, at which success stays at 50\% on queries between demonstration frames, the empirical counterpart of the radius $r_k$ of \cref{sec:analysis}. The aim--arrive gap is the $\ell_1$ distance between the upper level's predicted landing and the executed one.

\subsection{Language Goals}
\label{sec:goal}

The standard LIBERO and RoboCasa protocols specify tasks by language, while latent planning needs a latent goal. A light goal head maps the instruction and first latent to $J$ candidate final latents,
\begin{equation}
  \zgoal^{(j)} = \Gw\big(\mathrm{Enc}_{\mathrm{text}}(\ell), z_0, w_j\big), \quad
  \mathcal{L}_G = \min_j \big\| \zgoal^{(j)} - E(o_T) \big\|_2^2 ,
  \label{eq:goal}
\end{equation}
with noise $w_j \sim \mathcal{N}(0, I)$, trained on demonstration final frames with a frozen text encoder, following language-to-latent goal prediction~\cite{barbeau2026lago}. The $J$ candidates cover instructions with several valid final states, and \cref{eq:energy} scores each plan against the candidate nearest to its predicted landing. The instruction thus enters the planner only through the goal latents. With a goal image, $J = 1$ and $\zgoal = E(o_g)$.

\subsection{Planning Radius Analysis}
\label{sec:analysis}

We analyze RS-JEPA with deterministic dynamics and a perfect $\Pz$, which isolates what recursion and executed training each provide. The planning radius $r_k$ of $\pik{k}$ is the largest control distance, in steps, at which it succeeds with probability $1/2$. A self-model is \emph{calibrated} when it equals the median landing of \cref{eq:selfmodel}, and search is \emph{exact} when it returns a minimizer of \cref{eq:energy} over at most $M$ subgoals, breaking ties toward the shortest plan. Supplementary Sec.~A gives the formal statements and proofs.

\paragraph{Metric Limit of Flat Planning.}
A flat planner of horizon $H$ that scores plans by latent distance stalls at any state where nothing within $H$ steps is closer to the goal. The limit holds with a perfect world model, in line with~\cite{alrasheed2026planning,singh2026objective}. A goal at small latent distance can stay out of reach, and closer goals, which the upper levels supply, remove the limit.

\paragraph{Collapse under Optimism.}
Write the self-model as its aim plus a gap, $\Pt(z, g, k) = g + \Delta_\theta(z, g, k)$. A rollout then ends at
\begin{equation}
  \zhat_M = g_M + \Delta_\theta(\zhat_{M-1}, g_M, K),
  \label{eq:gap}
\end{equation}
so earlier subgoals reach the energy only through the gap. With $\Delta_\theta \equiv 0$, the energy depends on the last subgoal alone, delegating the goal is optimal, and a hierarchy of any depth behaves as its flat planner. The pretrained self-model satisfies $\Delta_\theta \equiv 0$, and an upper level adds reach only through the gap, which executed outcomes supply.

\begin{theorem}[Recursive reach]
\label{thm:reach}
With calibrated self-models and exact search, a hierarchy of depth $K$ reaches every goal within control distance $r = H M^K$. With a search budget $(N, I)$ per upper level, a decision costs
\begin{equation}
  C_K \le K N I M + N_0 I_0 H ,
  \label{eq:cost}
\end{equation}
and when a budget that does not depend on depth realizes the exact search, the hierarchy reaches radius $r$ at this cost. There are environments in which a flat planner that scores by latent distance reaches these goals only with horizon $r$, at a cost of at least $N_f I_f\, r$.
\end{theorem}

Radius thus grows exponentially with depth while compute grows linearly, and at equal samples per search the compute gap to flat search widens as $\Omega(r / \log r)$. With stochastic landings, missed subgoals are retried. If each subgoal advances a fraction $\eta$ of the radius below, then $r_K \ge r_0 (\eta M)^K$, and self-improvement raises $\eta$.

\paragraph{Critical Dimension of Search.}
CEM searches a plan of $M$ codes, a point in dimension $M d_u$. When the prior mean misses the optimal plan by an error on the scale of its width $\sq$, samples concentrate at distance $\sq \sqrt{2 M d_u}$ from the optimum. If the energy carries signal within distance $\rho$ of the optimum, search receives signal only while $M d_u$ stays below the critical dimension
\begin{equation}
  d_c = \frac{\rho^2}{2 \sq^2} ,
  \label{eq:critical}
\end{equation}
and above it the refit stops moving toward the optimum. Each $\sqrt{2}$ reduction of $\sq$ doubles $d_c$, so a learned prior and a low-dimensional code work together, as in the compressed high-level actions of~\cite{zhang2026hwm,caselli2026hilewm}.

\begin{theorem}[Grounded self-improvement]
\label{thm:fixed}
Train self-models from a class $\Fc$ on executed outcomes. While the levels below stay fixed, a self-model consistent with its data errs by more than $\alpha$ on at most $\dimE(\Fc, \alpha)$ of the queries that search selects, for any search rule. With executed labels at every input and a pointwise $\ell_1$ fit, which resolves conflicting hindsight labels, rounds settle one level each and reach the calibrated hierarchy, the unique fixed point, within $K$ rounds from any initialization, and starting from optimism the planner after round $t$ reaches every goal within control distance $H M^t$. Trained on labels imagined by the self-model itself, optimism remains a fixed point.
\end{theorem}

Here $\dimE$ is the eluder dimension~\cite{russo2013eluder}. Selecting the best of $N$ candidates under prediction noise $\sigma$ biases the selected prediction by up to $\sigma \sqrt{2 \ln N}$, which imagined labels carry from round to round, and the first claim bounds how often search can still select a mispredicted query.

\section{Experiments}
\label{sec:experiments}

\subsection{Setup}
\label{sec:setup}

\paragraph{Benchmarks.}
We evaluate on the four LIBERO suites~\cite{liu2023libero} and the standard RoboCasa task set~\cite{nasiriany2024robocasa}, both with simulated Franka arms, under their language-conditioned protocols, with goal-image results alongside. Evaluation uses each benchmark's fixed initial states and scenes, and self-improvement samples from a disjoint pool for $R = 5$ rounds of about 2{,}000 episodes each. Supplementary Sec.~C gives the full protocol.

\paragraph{Baselines.}
We compare planners that act through a latent world model; vision-language-action policies~\cite{kim2024openvla,black2025pi0} form a separate family. The baselines are V-JEPA 2-AC~\cite{assran2025vjepa2} with flat CEM, the hierarchical planners HWM~\cite{zhang2026hwm} and H-JEPA~\cite{zhang2026hjepa}, the subgoal generator SAGE~\cite{cheng2026sage}, the retrieval-based Metro-WM~\cite{lee2026metro}, the amortized planner RP1~\cite{sommer2026rp1}, and, for efficiency, V-JEPA 2-AC with the reachability loss of RC-aux~\cite{li2026rcaux}. All planners share the encoder and goal head at their lowest level, and V-JEPA 2-AC tunes its horizon and sample count on validation initial states. No baseline reports results on these suites, so all run under one protocol and one interaction budget, which RS-JEPA spends on self-improvement. Each baseline fine-tunes its world models with one Dyna pass~\cite{sutton1991dyna} per round, SAGE also retrains its subgoal generator with hindsight, and Metro-WM adds executed frames to its graph, so every upper level trains on its own episodes within the same budget.

\paragraph{Metrics.}
We report success rate (SR) over 3 seeds, $\rfifty$, the aim--arrive gap, and model evaluations per decision. A decision is one Level-0 replan, a model evaluation is one candidate advanced one step by $\Pz$ or $\Pt$, and each upper-level search is spread over the decisions it serves.

\subsection{Manipulation Benchmarks}
\label{sec:main}

\begin{table*}[t]
  \centering
  \caption{\textbf{Manipulation benchmarks.} Success rate (\%), mean $\pm$ standard deviation over 3 seeds, on the V-JEPA 2-AC backbone with a shared interaction budget. The indented RS-JEPA controls either plan with an upper-level world model or use the decoded prior mean as the subgoal (\cref{tab:ablation}).}
  \label{tab:main}
  \small
  \setlength{\tabcolsep}{2.7pt}
  \begin{tabular}{lcccccc}
    \toprule
    & \multicolumn{5}{c}{Language goal (\%)} & Image goal (\%) \\
    \cmidrule(lr){2-6} \cmidrule(lr){7-7}
    Method & LIBERO-Spatial $\uparrow$ & LIBERO-Object $\uparrow$ & LIBERO-Goal $\uparrow$ & LIBERO-Long $\uparrow$ & RoboCasa $\uparrow$ & LIBERO-Long $\uparrow$ \\
    \midrule
    V-JEPA 2-AC~\cite{assran2025vjepa2} & 48.4{\scriptsize\,\ensuremath{\pm}\,3.3} & 59.9{\scriptsize\,\ensuremath{\pm}\,3.3} & 42.7{\scriptsize\,\ensuremath{\pm}\,2.7} & 11.6{\scriptsize\,\ensuremath{\pm}\,2.1} & 19.3{\scriptsize\,\ensuremath{\pm}\,2.1} & 13.1{\scriptsize\,\ensuremath{\pm}\,2.6} \\
    HWM~\cite{zhang2026hwm} & 54.3{\scriptsize\,\ensuremath{\pm}\,2.0} & 60.6{\scriptsize\,\ensuremath{\pm}\,2.5} & 49.0{\scriptsize\,\ensuremath{\pm}\,3.6} & 20.3{\scriptsize\,\ensuremath{\pm}\,2.9} & 21.4{\scriptsize\,\ensuremath{\pm}\,1.6} & 21.2{\scriptsize\,\ensuremath{\pm}\,2.2} \\
    H-JEPA~\cite{zhang2026hjepa} & 51.7{\scriptsize\,\ensuremath{\pm}\,3.1} & 56.9{\scriptsize\,\ensuremath{\pm}\,2.6} & 48.3{\scriptsize\,\ensuremath{\pm}\,2.1} & 24.0{\scriptsize\,\ensuremath{\pm}\,1.7} & 25.8{\scriptsize\,\ensuremath{\pm}\,2.3} & 24.5{\scriptsize\,\ensuremath{\pm}\,2.6} \\
    SAGE~\cite{cheng2026sage} & 51.6{\scriptsize\,\ensuremath{\pm}\,2.4} & 55.8{\scriptsize\,\ensuremath{\pm}\,2.7} & 47.7{\scriptsize\,\ensuremath{\pm}\,2.1} & 30.8{\scriptsize\,\ensuremath{\pm}\,3.2} & 27.2{\scriptsize\,\ensuremath{\pm}\,2.3} & 32.6{\scriptsize\,\ensuremath{\pm}\,2.2} \\
    Metro-WM~\cite{lee2026metro} & 57.0{\scriptsize\,\ensuremath{\pm}\,1.6} & \textbf{63.4}{\scriptsize\,\ensuremath{\pm}\,3.3} & \underline{50.4}{\scriptsize\,\ensuremath{\pm}\,3.0} & 37.2{\scriptsize\,\ensuremath{\pm}\,3.5} & \underline{29.0}{\scriptsize\,\ensuremath{\pm}\,2.0} & 36.5{\scriptsize\,\ensuremath{\pm}\,3.1} \\
    RP1~\cite{sommer2026rp1} & 47.1{\scriptsize\,\ensuremath{\pm}\,1.6} & 60.7{\scriptsize\,\ensuremath{\pm}\,1.8} & 42.6{\scriptsize\,\ensuremath{\pm}\,3.8} & 12.5{\scriptsize\,\ensuremath{\pm}\,2.6} & 19.5{\scriptsize\,\ensuremath{\pm}\,1.3} & 14.3{\scriptsize\,\ensuremath{\pm}\,1.9} \\
    \midrule
    RS-JEPA (Ours) & \textbf{63.9}{\scriptsize\,\ensuremath{\pm}\,2.2} & \underline{62.3}{\scriptsize\,\ensuremath{\pm}\,3.5} & \textbf{61.0}{\scriptsize\,\ensuremath{\pm}\,3.1} & \textbf{50.7}{\scriptsize\,\ensuremath{\pm}\,2.4} & \textbf{32.6}{\scriptsize\,\ensuremath{\pm}\,2.4} & \textbf{51.9}{\scriptsize\,\ensuremath{\pm}\,1.6} \\
    \quad World-model upper level & 55.6{\scriptsize\,\ensuremath{\pm}\,1.7} & 60.5{\scriptsize\,\ensuremath{\pm}\,1.4} & 49.6{\scriptsize\,\ensuremath{\pm}\,3.5} & 27.4{\scriptsize\,\ensuremath{\pm}\,1.9} & 26.5{\scriptsize\,\ensuremath{\pm}\,2.5} & 28.3{\scriptsize\,\ensuremath{\pm}\,3.9} \\
    \quad Prior-mean subgoal & \underline{58.1}{\scriptsize\,\ensuremath{\pm}\,2.9} & 61.6{\scriptsize\,\ensuremath{\pm}\,3.1} & 50.2{\scriptsize\,\ensuremath{\pm}\,2.4} & \underline{38.6}{\scriptsize\,\ensuremath{\pm}\,3.6} & 27.2{\scriptsize\,\ensuremath{\pm}\,1.9} & \underline{40.3}{\scriptsize\,\ensuremath{\pm}\,3.6} \\
    \bottomrule
  \end{tabular}
\end{table*}

\begin{figure}[t]
  \centering
  \includegraphics[width=\linewidth]{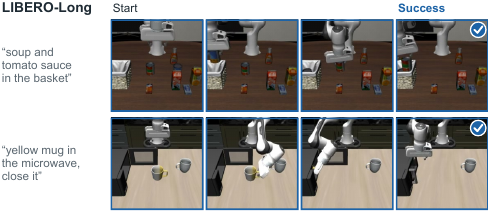}
  \caption{\textbf{Qualitative results on LIBERO-Long.} One RS-JEPA episode for each of two tasks, with four frames from the start to success; quotes abbreviate the task instructions. Supplementary Sec.~E shows all four suites.}
  \label{fig:qualmain}
\end{figure}

On LIBERO-Long (\cref{tab:main}), RS-JEPA reaches 50.7\%, against 37.2\% for Metro-WM, the strongest baseline. \Cref{fig:qualmain} shows two of its episodes. LIBERO-Object tasks need little decomposition, and Metro-WM is slightly higher there (63.4\% vs.\ 62.3\%). With the same codes, prior and executed rounds, both controls stay below RS-JEPA on every suite. With equal executed data, predicting the planner's landing therefore outperforms predicting the world or proposing from a hindsight prior.

Hierarchical baselines gain most on LIBERO-Long, while H-JEPA and SAGE trail flat search on LIBERO-Object. Metro-WM comes closest to RS-JEPA. Its retrieved frames are realizable states, yet such a state can lie beyond the planner's reach from where it stands. RP1 reduces search cost at a fixed horizon and stays close to V-JEPA 2-AC.

The image-goal column tracks the language-goal column, so the goal head adds little error, and the same order holds on the LeWM backbone (supplementary Sec.~E).

\subsection{Planning Radius and Compute}
\label{sec:radius}

\begin{figure*}[t]
  \centering
  \includegraphics[width=\linewidth]{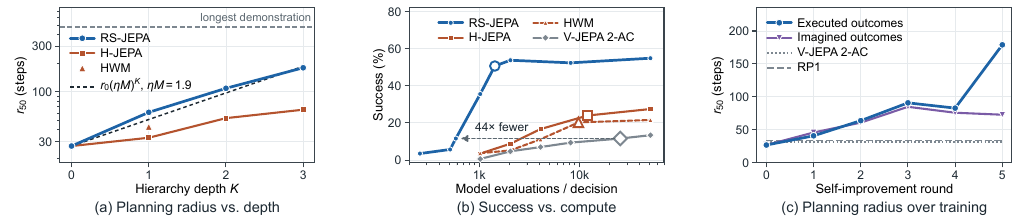}
  \caption{\textbf{Planning radius and compute} on LIBERO-Long. (a) $\rfifty$ against depth, which grows geometrically as in \cref{thm:reach} and stays below the longest demonstration. (b) Success against model evaluations per decision; open markers are the operating points of \cref{tab:efficiency}, and the arrow marks where RS-JEPA matches V-JEPA 2-AC. (c) $\rfifty$ over self-improvement rounds with executed or imagined outcomes.}
  \label{fig:radius}
\end{figure*}

\begin{table}[t]
  \centering
  \caption{\textbf{Planning efficiency} on LIBERO-Long. Evaluations count every candidate step of $\Pz$ and $\Pt$; time is per decision.}
  \label{tab:efficiency}
  \small
  \setlength{\tabcolsep}{2.5pt}
  \begin{tabular}{lcccc}
    \toprule
    Method & SR (\%) $\uparrow$ & \shortstack{$\rfifty$\\(steps) $\uparrow$} & \shortstack{Model evals per\\decision (k) $\downarrow$} & \shortstack{Time\\(ms) $\downarrow$} \\
    \midrule
    V-JEPA 2-AC~\cite{assran2025vjepa2} & 11.6 & 32 & 25.60 & 1591 \\
    \quad + RC-aux~\cite{li2026rcaux} & 13.3 & 44 & 28.26 & 1748 \\
    HWM~\cite{zhang2026hwm} & 20.3 & 43 & 9.80 & 636 \\
    H-JEPA~\cite{zhang2026hjepa} & 24.0 & 65 & 12.09 & 781 \\
    Metro-WM~\cite{lee2026metro} & \underline{37.2} & \underline{90} & 3.07 & 210 \\
    RP1~\cite{sommer2026rp1} & 12.5 & 34 & \textbf{0.45} & \textbf{48} \\
    \midrule
    RS-JEPA (Ours) & \textbf{50.7} & \textbf{179} & \underline{1.43} & \underline{121} \\
    \midrule
    \multicolumn{5}{l}{\emph{Reference}} \\
    Oracle subgoals & 61.2 & 221 & 1.15 & 75 \\
    \bottomrule
  \end{tabular}
\end{table}

Each level multiplies $\rfifty$ by about 1.9 (\cref{fig:radius}(a)), the factor $\eta M$ of \cref{sec:analysis} with three subgoals per plan. Growth slows at higher levels, as longer pursuits miss more often. Among planners that search at every level (\cref{fig:radius}(b)), RS-JEPA needs about 44$\times$ fewer model evaluations per decision than V-JEPA 2-AC at equal success, and about 18$\times$ fewer at the operating points of \cref{tab:efficiency}. HWM and H-JEPA lie between the two. Temporal decomposition saves compute, and the learned prior and small search widen the gap. Supplementary Sec.~E illustrates the search strategies on one LIBERO-Long query.

In \cref{tab:efficiency}, RS-JEPA reaches the longest radius with fewer evaluations per decision than every method except RP1, and self-improvement takes about 200 GPU-hours per suite. RC-aux extends the flat radius but still searches the full horizon at every decision. Metro-WM keeps a large Level-0 search because retrieved frames often lie beyond a small one. RP1 amortizes its search at a far shorter radius.

Demonstration latents as subgoals on the same Level 0 reach 61.2\%, and RS-JEPA comes within 10.5 points while proposing every subgoal itself. Supplementary Sec.~E sweeps the code dimension. Success falls sharply near the dimension predicted by \cref{eq:critical}, and self-improvement moves this point to larger $d_u$.

\subsection{Self-Improvement}
\label{sec:improve}

Over rounds (\cref{fig:radius}(c)), V-JEPA 2-AC and RP1 stay level, since the flat limit persists under a perfect world model (\cref{sec:analysis}). RS-JEPA extends its radius from 27 to 179 steps, with a dip at round 4.

Level-0 search selects the actions on which $\Pz$ errs, so landings imagined by $\Pz$ overstate arrival, and the selection bias of \cref{sec:analysis} enters the labels. With imagined outcomes, the radius rises while subgoals stay short, then falls to 73 steps as the bias accumulates over rounds. Executed training keeps the aim--arrive gap small (\cref{tab:ablation}), since each round's landings correct the subgoals search favored.

\subsection{Ablation Study}
\label{sec:ablation}

\begin{table}[t]
  \centering
  \caption{\textbf{Ablation study.} Gap and evaluations per decision on LIBERO-Long. The Gaussian-prior variant reaches full-model success.}
  \label{tab:ablation}
  \footnotesize
  \setlength{\tabcolsep}{2.5pt}
  \begin{tabular}{lcccc}
    \toprule
    & \multicolumn{2}{c}{SR (\%)} & & \\
    \cmidrule(lr){2-3}
    Variant & \shortstack{LIBERO-\\Long $\uparrow$} & \shortstack{LIBERO-\\Goal $\uparrow$} & Gap $\downarrow$ & \shortstack{Evals\\(k) $\downarrow$} \\
    \midrule
    \textbf{Full model} & \textbf{50.7} & \underline{61.0} & \textbf{0.20} & 1.432 \\
    \midrule
    \multicolumn{5}{l}{\emph{Upper level}} \\
    Self-model $\to$ world model & 27.4 & 49.6 & 0.41 & 1.615 \\
    Self-model $\to$ prior mean & 38.6 & 50.2 & 0.30 & \textbf{1.178} \\
    Learned prior $\to$ Gaussian & \underline{50.4} & 60.6 & \textbf{0.20} & 3.902 \\
    \midrule
    \multicolumn{5}{l}{\emph{Training data}} \\
    w/o self-improvement (round 0) & 11.5 & 40.6 & 0.50 & 25.797 \\
    Executed $\to$ imagined outcomes & 31.3 & 52.9 & 0.46 & 1.448 \\
    w/o hindsight relabeling & 46.9 & 47.3 & 0.28 & 1.459 \\
    Frozen $\Pz$ $\to$ Dyna on $\Pz$ & 50.1 & \textbf{62.1} & \underline{0.21} & \underline{1.426} \\
    \bottomrule
  \end{tabular}
\end{table}

In \cref{tab:ablation}, an HWM-style world-model upper level with its own learned prior, the same search and the same round budget falls to 27.4\% on LIBERO-Long. Its predictor maps a latent and a macro-action code to the next latent, so it models executed action chunks and not the pursuit of a subgoal. Subgoals the lower level misses therefore still look reachable, the mismatch Hi-LeWM reports, and the aim--arrive gap doubles. The learned prior and the executed rounds lift this variant above HWM in \cref{tab:main}, and the self-model adds the larger part of the gain. Handing down the decoded prior mean without scoring or search reaches 38.6\%. The prior proposes latents the planner has reached, and it is the self-model that selects those leading to the goal. Imagined outcomes cost the most on LIBERO-Long. Hindsight matters most on LIBERO-Goal, where failed attempts still land near other valid goals.

The learned prior accounts for much of the compute saving. With Gaussian sampling at the same success, the planner needs 3902 evaluations per decision instead of 1432, still fewer than HWM and H-JEPA. Dyna on $\Pz$ slightly improves LIBERO-Goal and slightly lowers LIBERO-Long success. Without self-improvement and at the flat planner's budget, the optimistic pretrained self-model reaches 11.5\%, close to 11.6\% for V-JEPA 2-AC, as the collapse result of \cref{sec:analysis} predicts.

\subsection{Real Robot Experiments}
\label{sec:robot}

\begin{table}[t]
  \centering
  \caption{\textbf{Real robot.} Successes in 20 trials per task on a B601-RS arm with image goals. Round 0 precedes self-improvement.}
  \label{tab:robot}
  \footnotesize
  \setlength{\tabcolsep}{1.7pt}
  \begin{tabular}{lcccccc}
    \toprule
    & Short & \multicolumn{2}{c}{Detour} & \multicolumn{2}{c}{Long horizon} & Adaptation \\
    \cmidrule(lr){2-2} \cmidrule(lr){3-4} \cmidrule(lr){5-6} \cmidrule(lr){7-7}
    Method & \shortstack{Stack\\cups} & \shortstack{Blocked\\drawer} & \shortstack{Object in\\drawer} & \shortstack{Clean\\table} & \shortstack{Rings\\on peg} & \shortstack{Peg\\insertion} \\
    \midrule
    V-JEPA 2-AC~\cite{assran2025vjepa2} & 13/20 & 3/20 & 1/20 & 2/20 & 2/20 & 3/20 \\
    HWM~\cite{zhang2026hwm} & \underline{14}/20 & 4/20 & 2/20 & 4/20 & 3/20 & 3/20 \\
    H-JEPA~\cite{zhang2026hjepa} & \underline{14}/20 & 3/20 & 4/20 & 5/20 & 5/20 & \underline{5}/20 \\
    Metro-WM~\cite{lee2026metro} & \textbf{15}/20 & \underline{7}/20 & \underline{5}/20 & \underline{7}/20 & \underline{6}/20 & 3/20 \\
    \midrule
    RS-JEPA, round 0 & 13/20 & 3/20 & 1/20 & 3/20 & 2/20 & 4/20 \\
    RS-JEPA (Ours) & \underline{14}/20 & \textbf{12}/20 & \textbf{9}/20 & \textbf{8}/20 & \textbf{10}/20 & \textbf{11}/20 \\
    \bottomrule
  \end{tabular}
\end{table}

We use a B601-RS arm with an external and a wrist camera, with goals given as images. Six tasks each probe one property (\cref{tab:robot}): \emph{stack cups} is a short-horizon control; in \emph{blocked drawer} a cup must be moved before the drawer opens, and in \emph{object in drawer} a closed drawer must be opened before the cube is taken out, detours that latent distance misjudges; \emph{clean table} and \emph{rings on peg} chain stages in a fixed order; and \emph{peg insertion} tests self-adaptation after the Level-0 action limit is halved, which leaves the dynamics unchanged, with two practice rounds for every method. One set of models serves all six tasks. Every planner starts from the same V-JEPA 2-AC world model, fine-tuned from its DROID checkpoint on 180 teleoperated demonstrations, and then spends 360 robot episodes, RS-JEPA on self-improvement and each baseline on Dyna (supplementary Sec.~C).

Metro-WM is slightly ahead of RS-JEPA on cup stacking. V-JEPA 2-AC stalls on the blocked drawer, succeeding in 3 of 20 trials, and HWM pulls against the cup (\cref{fig:teaser}). RS-JEPA first moves the cup aside and succeeds in 12. With the cube inside the closed drawer, RS-JEPA opens the drawer first and succeeds in 9 trials. On the long-horizon tasks, RS-JEPA completes the rings in 10 trials, against 6 for Metro-WM, whose retrieved subgoals do not encode the order of the rings. After the action limit is halved, a once-reachable subgoal can lie beyond the weakened lower level, and the self-model learns this from the practice landings. RS-JEPA completes 11 peg insertions, against 3 for Metro-WM, whose world model the change leaves untouched. Without self-improvement, RS-JEPA succeeds in 26 of 120 trials, close to 24 for V-JEPA 2-AC, and the rounds on the robot raise this to 64 (see supplementary Sec.~E).

\section{Conclusion}
\label{sec:conclusion}

We proposed RS-JEPA, a recursive planner whose upper levels predict where the lower-level planner lands and train only on executed outcomes. It chooses reachable stepping stones and extends its reach over training. The same construction applies to any planner that hands subgoals to a lower-level controller.

\paragraph{Limitations.} The analysis covers deterministic dynamics, a perfect world model and exact search. The self-model predicts a median landing. Self-improvement uses about 10{,}000 episodes per suite, and language goals map to candidate final latents.

{
  \small
  \bibliographystyle{ieeenat_fullname}
  \bibliography{references}

@misc{assran2025vjepa2,
  title         = {{V-JEPA} 2: Self-Supervised Video Models Enable Understanding, Prediction and Planning},
  author        = {Assran, Mido and Bardes, Adrien and Fan, David and Garrido, Quentin and Howes, Russell and Komeili, Mojtaba and Muckley, Matthew and Rizvi, Ammar and Roberts, Claire and Sinha, Koustuv and Zholus, Artem and Arnaud, Sergio and Gejji, Abha and Martin, Ada and Hogan, Francois Robert and Dugas, Daniel and Bojanowski, Piotr and Khalidov, Vasil and Labatut, Patrick and Massa, Francisco and Szafraniec, Marc and Krishnakumar, Kapil and Li, Yong and Ma, Xiaodong and Chandar, Sarath and Meier, Franziska and LeCun, Yann and Rabbat, Michael and Ballas, Nicolas},
  year          = {2025},
  eprint        = {2506.09985},
  archiveprefix = {arXiv}
}

@inproceedings{zhou2025dinowm,
  title         = {{DINO-WM}: World Models on Pre-trained Visual Features enable Zero-shot Planning},
  author        = {Zhou, Gaoyue and Pan, Hengkai and LeCun, Yann and Pinto, Lerrel},
  booktitle     = {International Conference on Machine Learning},
  pages         = {79115--79135},
  year          = {2025},
  eprint        = {2411.04983},
  archiveprefix = {arXiv}
}

@inproceedings{sobal2025pldm,
  title         = {Learning from Reward-Free Offline Data: A Case for Planning with Latent Dynamics Models},
  author        = {Sobal, Vlad and Zhang, Wancong and Cho, Kyunghyun and Balestriero, Randall and Rudner, Tim G. J. and LeCun, Yann},
  booktitle     = {Advances in Neural Information Processing Systems},
  year          = {2025},
  eprint        = {2502.14819},
  archiveprefix = {arXiv}
}

@misc{maes2026lewm,
  title         = {{LeWorldModel}: Stable End-to-End Joint-Embedding Predictive Architecture from Pixels},
  author        = {Maes, Lucas and Le Lidec, Quentin and Scieur, Damien and LeCun, Yann and Balestriero, Randall},
  year          = {2026},
  eprint        = {2603.19312},
  archiveprefix = {arXiv}
}

@article{terver2026jepawm,
  title         = {What Drives Success in Physical Planning with Joint-Embedding Predictive World Models?},
  author        = {Terver, Basile and Yang, Tsung-Yen and Ponce, Jean and Bardes, Adrien and LeCun, Yann},
  journal       = {Transactions on Machine Learning Research},
  year          = {2026},
  eprint        = {2512.24497},
  archiveprefix = {arXiv}
}

@misc{alrasheed2026planning,
  title         = {The Planning Limits of Latent World Models},
  author        = {Alrasheed, Ali and Azam, Basim and Akhtar, Naveed},
  year          = {2026},
  eprint        = {2609.39235},
  archiveprefix = {arXiv}
}

@misc{singh2026objective,
  title         = {The Objective Is the Bottleneck: Latent World Models Encode What Their Planners Cannot Use},
  author        = {Singh, Joyjeet},
  year          = {2026},
  eprint        = {2608.12959},
  archiveprefix = {arXiv}
}

@misc{gao2026imwm,
  title         = {{IMWM}: Intuition Models Complement World Models for Latent Planning},
  author        = {Gao, Baoqi and Han, Ruize and Wang, Miao and Wang, Song},
  year          = {2026},
  eprint        = {2606.01626},
  archiveprefix = {arXiv}
}

@inproceedings{zhang2026hwm,
  title         = {Hierarchical Planning with Latent World Models},
  author        = {Zhang, Wancong and Terver, Basile and Zholus, Artem and Chitnis, Soham and Sutaria, Harsh and Assran, Mido and Balestriero, Randall and Bar, Amir and LeCun, Yann and Ballas, Nicolas},
  booktitle     = {Conference on Robot Learning},
  year          = {2026},
  eprint        = {2604.03208},
  archiveprefix = {arXiv}
}

@misc{zhang2026hjepa,
  title         = {{H-JEPA}: End-to-End Learning of Hierarchical World Models for Visual Planning},
  author        = {Zhang, Wancong and Terver, Basile and Rabbat, Michael and LeCun, Yann and Balestriero, Randall},
  year          = {2026},
  eprint        = {2610.06805},
  archiveprefix = {arXiv}
}

@inproceedings{caselli2026hilewm,
  title         = {Mind the Gap: Promises and Pitfalls of Hierarchical Planning in {LeWorldModel}},
  author        = {Caselli, Niccol{\`o} and Massafra, Francesco and Punzo, Samuele and Lo Sardo, Salvatore and Pantelidis, Ippokratis and Bhethanabhotla, Sathya Kamesh},
  booktitle     = {WM@Booth Workshop},
  year          = {2026},
  eprint        = {2607.12547},
  archiveprefix = {arXiv}
}

@misc{masip2026ffjepa,
  title         = {{FF-JEPA}: Long-Horizon Planning in World Models with Latent Planners},
  author        = {Masip, Sergi and Swinnen, Jonathan and Hu, Yutong and Detry, Renaud and Tuytelaars, Tinne},
  year          = {2026},
  eprint        = {2606.09311},
  archiveprefix = {arXiv}
}

@misc{barbeau2026lago,
  title         = {Latent Goal Prediction from Language for Model-Based Planning},
  author        = {Barbeau, Samuel and Roy, Simon and Beltrame, Giovanni and Desrosiers, Christian and Thome, Nicolas},
  year          = {2026},
  eprint        = {2606.20627},
  archiveprefix = {arXiv}
}

@misc{cheng2026sage,
  title         = {{SAGE}: Subgoal-Conditioned Action Generation for Latent World Model Planning},
  author        = {Cheng, Letian and Zhang, Qi and Wang, Yisen},
  year          = {2026},
  eprint        = {2607.17973},
  archiveprefix = {arXiv}
}

@misc{xiang2026frozen,
  title         = {Learning Goal-Conditioned Subgoals through Frozen Hierarchical World Models},
  author        = {Xiang, Kaibo and Liu, Qingfeng},
  year          = {2026},
  eprint        = {2610.26179},
  archiveprefix = {arXiv}
}

@misc{nguyen2026gcidm,
  title         = {Latent Geometry Beyond Search: Amortizing Planning in World Models},
  author        = {Nguyen, Hoang and Xu, Xiaohao and Huang, Xiaonan},
  year          = {2026},
  eprint        = {2605.08732},
  archiveprefix = {arXiv}
}

@misc{huang2026leflow,
  title         = {{LeFlow}: Generative Latent Flow Planning for World Models},
  author        = {Huang, Hsiang-Wei and Shangguan, Jianxu and Lu, Junbin and Hwang, Jenq-Neng},
  year          = {2026},
  eprint        = {2608.24855},
  archiveprefix = {arXiv}
}

@misc{bansal2026leplanner,
  title         = {{LePlanner}: An Iterative Amortized Controller For World Models},
  author        = {Bansal, Saksham and Naphade, Om and Aggarwal, Chayan and M, Vrishin},
  year          = {2026},
  eprint        = {2609.13845},
  archiveprefix = {arXiv}
}

@misc{sommer2026rp1,
  title         = {Reinforced Planning with Latent World Models},
  author        = {Sommer, Armin and Schilling, Jannik},
  year          = {2026},
  eprint        = {2608.18669},
  archiveprefix = {arXiv}
}

@inproceedings{shi2022skimo,
  title         = {Skill-based Model-based Reinforcement Learning},
  author        = {Shi, Lucy Xiaoyang and Lim, Joseph J. and Lee, Youngwoon},
  booktitle     = {Conference on Robot Learning},
  year          = {2022},
  eprint        = {2207.07560},
  archiveprefix = {arXiv}
}

@inproceedings{sharma2020dads,
  title         = {Dynamics-Aware Unsupervised Discovery of Skills},
  author        = {Sharma, Archit and Gu, Shixiang and Levine, Sergey and Kumar, Vikash and Hausman, Karol},
  booktitle     = {International Conference on Learning Representations},
  year          = {2020},
  eprint        = {1907.01657},
  archiveprefix = {arXiv}
}

@inproceedings{nasiriany2019leap,
  title         = {Planning with Goal-Conditioned Policies},
  author        = {Nasiriany, Soroush and Pong, Vitchyr H. and Lin, Steven and Levine, Sergey},
  booktitle     = {Advances in Neural Information Processing Systems},
  year          = {2019},
  eprint        = {1911.08453},
  archiveprefix = {arXiv}
}

@inproceedings{li2026rcaux,
  title         = {Predictive but Not Plannable: {RC-aux} for Latent World Models},
  author        = {Li, Wenyuan and Li, Guang and Maeda, Keisuke and Ogawa, Takahiro and Haseyama, Miki},
  booktitle     = {Advances in Neural Information Processing Systems},
  year          = {2026},
  eprint        = {2605.07278},
  archiveprefix = {arXiv}
}

@inproceedings{levy2026simdist,
  title         = {Simulation Distillation: Pretraining World Models in Simulation for Rapid Real-World Adaptation},
  author        = {Levy, Jacob and Westenbroek, Tyler and Huang, Kevin and Palafox, Fernando and Yin, Patrick and Omidshafiei, Shayegan and Kim, Dong-Ki and Gupta, Abhishek and Fridovich-Keil, David},
  booktitle     = {Robotics: Science and Systems},
  year          = {2026},
  eprint        = {2603.15759},
  archiveprefix = {arXiv}
}

@inproceedings{yang2026rise,
  title         = {{RISE}: Self-Improving Robot Policy with Compositional World Model},
  author        = {Yang, Jiazhi and Lin, Kunyang and Li, Jinwei and Zhang, Wencong and Lin, Tianwei and Wu, Longyan and Su, Zhizhong and Zhao, Hao and Zhang, Ya-Qin and Chen, Li and Luo, Ping and Yue, Xiangyu and Li, Hongyang},
  booktitle     = {Robotics: Science and Systems},
  year          = {2026},
  eprint        = {2602.11075},
  archiveprefix = {arXiv}
}

@misc{bhamidipaty2026exploitable,
  title         = {Imperfect World Models are Exploitable},
  author        = {Bhamidipaty, Logan Mondal and Whitammer, Esmeralda S. and Abel, David and Kochenderfer, Mykel J. and Ramamoorthy, Subramanian},
  year          = {2026},
  eprint        = {2605.15960},
  archiveprefix = {arXiv}
}

@inproceedings{andrychowicz2017her,
  title         = {Hindsight Experience Replay},
  author        = {Andrychowicz, Marcin and Wolski, Filip and Ray, Alex and Schneider, Jonas and Fong, Rachel and Welinder, Peter and McGrew, Bob and Tobin, Josh and Abbeel, Pieter and Zaremba, Wojciech},
  booktitle     = {Advances in Neural Information Processing Systems},
  year          = {2017},
  eprint        = {1707.01495},
  archiveprefix = {arXiv}
}

@inproceedings{russo2013eluder,
  title         = {Eluder Dimension and the Sample Complexity of Optimistic Exploration},
  author        = {Russo, Daniel and Van Roy, Benjamin},
  booktitle     = {Advances in Neural Information Processing Systems},
  volume        = {26},
  pages         = {2256--2264},
  year          = {2013}
}

@article{sutton1991dyna,
  title         = {Dyna, an Integrated Architecture for Learning, Planning, and Reacting},
  author        = {Sutton, Richard S.},
  journal       = {ACM SIGART Bulletin},
  volume        = {2},
  number        = {4},
  pages         = {160--163},
  year          = {1991},
  doi           = {10.1145/122344.122377}
}

@inproceedings{liu2023libero,
  title         = {{LIBERO}: Benchmarking Knowledge Transfer for Lifelong Robot Learning},
  author        = {Liu, Bo and Zhu, Yifeng and Gao, Chongkai and Feng, Yihao and Liu, Qiang and Zhu, Yuke and Stone, Peter},
  booktitle     = {Advances in Neural Information Processing Systems},
  year          = {2023},
  eprint        = {2306.03310},
  archiveprefix = {arXiv}
}

@inproceedings{nasiriany2024robocasa,
  title         = {{RoboCasa}: Large-Scale Simulation of Everyday Tasks for Generalist Robots},
  author        = {Nasiriany, Soroush and Maddukuri, Abhiram and Zhang, Lance and Parikh, Adeet and Lo, Aaron and Joshi, Abhishek and Mandlekar, Ajay and Zhu, Yuke},
  booktitle     = {Robotics: Science and Systems},
  year          = {2024},
  eprint        = {2406.02523},
  archiveprefix = {arXiv}
}

@article{hafner2025dreamer,
  title   = {Mastering Diverse Control Tasks through World Models},
  author  = {Hafner, Danijar and Pasukonis, Jurgis and Ba, Jimmy and Lillicrap, Timothy},
  journal = {Nature},
  volume  = {640},
  number  = {8059},
  pages   = {647--653},
  year    = {2025},
  doi     = {10.1038/s41586-025-08744-2}
}

@inproceedings{hansen2024tdmpc2,
  title         = {{TD-MPC2}: Scalable, Robust World Models for Continuous Control},
  author        = {Hansen, Nicklas and Su, Hao and Wang, Xiaolong},
  booktitle     = {International Conference on Learning Representations},
  year          = {2024},
  eprint        = {2310.16828},
  archiveprefix = {arXiv}
}

@inproceedings{bar2025nwm,
  title         = {Navigation World Models},
  author        = {Bar, Amir and Zhou, Gaoyue and Tran, Danny and Darrell, Trevor and LeCun, Yann},
  booktitle     = {Proceedings of the IEEE/CVF Conference on Computer Vision and Pattern Recognition (CVPR)},
  year          = {2025},
  eprint        = {2412.03572},
  archiveprefix = {arXiv}
}

@article{rubinstein1999cem,
  title   = {The Cross-Entropy Method for Combinatorial and Continuous Optimization},
  author  = {Rubinstein, Reuven},
  journal = {Methodology and Computing in Applied Probability},
  volume  = {1},
  number  = {2},
  pages   = {127--190},
  year    = {1999},
  doi     = {10.1023/A:1010091220143}
}

@inproceedings{chua2018pets,
  title         = {Deep Reinforcement Learning in a Handful of Trials using Probabilistic Dynamics Models},
  author        = {Chua, Kurtland and Calandra, Roberto and McAllister, Rowan and Levine, Sergey},
  booktitle     = {Advances in Neural Information Processing Systems},
  year          = {2018},
  eprint        = {1805.12114},
  archiveprefix = {arXiv}
}

@inproceedings{hafner2022director,
  title         = {Deep Hierarchical Planning from Pixels},
  author        = {Hafner, Danijar and Lee, Kuang-Huei and Fischer, Ian and Abbeel, Pieter},
  booktitle     = {Advances in Neural Information Processing Systems},
  year          = {2022},
  eprint        = {2206.04114},
  archiveprefix = {arXiv}
}

@inproceedings{park2023hiql,
  title         = {{HIQL}: Offline Goal-Conditioned {RL} with Latent States as Actions},
  author        = {Park, Seohong and Ghosh, Dibya and Eysenbach, Benjamin and Levine, Sergey},
  booktitle     = {Advances in Neural Information Processing Systems},
  year          = {2023},
  eprint        = {2307.11949},
  archiveprefix = {arXiv}
}

@misc{lee2026metro,
  title         = {{Metro-WM}: Long-Horizon Latent Planning with Realisable Sub-Goals},
  author        = {Lee, Royson and Rezk, Fady and Parcollet, Titouan and Hospedales, Timothy and Cornelio, Cristina},
  year          = {2026},
  eprint        = {2609.31868},
  archiveprefix = {arXiv}
}

@misc{liu2026wav,
  title         = {World Action Verifier: Self-Improving World Models via Forward-Inverse Asymmetry},
  author        = {Liu, Yuejiang and Feng, Fan and Kong, Lingjing and Lu, Weifeng and Tang, Jinzhou and Zhang, Kun and Murphy, Kevin and Finn, Chelsea and Du, Yilun},
  year          = {2026},
  eprint        = {2604.01985},
  archiveprefix = {arXiv}
}

@misc{cui2026rsi,
  title         = {{RSI-Anything}: Unlocking Recursive Self-Improvement with World Models},
  author        = {Cui, Taoyong and Zeng, Pinxian and Xu, Wanghan and Xiao, Hao and Xue, Shuhan and Gao, Qiang and Ouyang, Wanli and Heng, Pheng Ann and Yin, Zhenfei and Wu, Yingcheng and Yang, Ling},
  year          = {2026},
  eprint        = {2610.13912},
  archiveprefix = {arXiv}
}

@article{bongard2006resilient,
  title   = {Resilient Machines Through Continuous Self-Modeling},
  author  = {Bongard, Josh and Zykov, Victor and Lipson, Hod},
  journal = {Science},
  volume  = {314},
  number  = {5802},
  pages   = {1118--1121},
  year    = {2006},
  doi     = {10.1126/science.1133687}
}

@article{chen2022selfmodel,
  title   = {Fully Body Visual Self-Modeling of Robot Morphologies},
  author  = {Chen, Boyuan and Kwiatkowski, Robert and Vondrick, Carl and Lipson, Hod},
  journal = {Science Robotics},
  volume  = {7},
  number  = {68},
  pages   = {eabn1944},
  year    = {2022},
  doi     = {10.1126/scirobotics.abn1944}
}

@inproceedings{janner2019mbpo,
  title         = {When to Trust Your Model: Model-Based Policy Optimization},
  author        = {Janner, Michael and Fu, Justin and Zhang, Marvin and Levine, Sergey},
  booktitle     = {Advances in Neural Information Processing Systems},
  year          = {2019},
  eprint        = {1906.08253},
  archiveprefix = {arXiv}
}

@inproceedings{lakshminarayanan2017ensembles,
  title         = {Simple and Scalable Predictive Uncertainty Estimation using Deep Ensembles},
  author        = {Lakshminarayanan, Balaji and Pritzel, Alexander and Blundell, Charles},
  booktitle     = {Advances in Neural Information Processing Systems},
  year          = {2017},
  eprint        = {1612.01474},
  archiveprefix = {arXiv}
}

@article{smith2006curse,
  title   = {The Optimizer's Curse: Skepticism and Postdecision Surprise in Decision Analysis},
  author  = {Smith, James E. and Winkler, Robert L.},
  journal = {Management Science},
  volume  = {52},
  number  = {3},
  pages   = {311--322},
  year    = {2006},
  doi     = {10.1287/mnsc.1050.0451}
}

@inproceedings{wu2023daydreamer,
  title         = {{DayDreamer}: World Models for Physical Robot Learning},
  author        = {Wu, Philipp and Escontrela, Alejandro and Hafner, Danijar and Abbeel, Pieter and Goldberg, Ken},
  booktitle     = {Conference on Robot Learning},
  volume        = {205},
  pages         = {2226--2240},
  year          = {2022},
  eprint        = {2206.14176},
  archiveprefix = {arXiv}
}

@inproceedings{khazatsky2024droid,
  title         = {{DROID}: A Large-Scale In-The-Wild Robot Manipulation Dataset},
  author        = {Khazatsky, Alexander and Pertsch, Karl and Nair, Suraj and others},
  booktitle     = {Robotics: Science and Systems},
  year          = {2024},
  eprint        = {2403.12945},
  archiveprefix = {arXiv}
}

@inproceedings{ghosh2021gcsl,
  title         = {Learning to Reach Goals via Iterated Supervised Learning},
  author        = {Ghosh, Dibya and Gupta, Abhishek and Reddy, Ashwin and Fu, Justin and Devin, Coline and Eysenbach, Benjamin and Levine, Sergey},
  booktitle     = {International Conference on Learning Representations},
  year          = {2021},
  eprint        = {1912.06088},
  archiveprefix = {arXiv}
}

@inproceedings{levy2019hac,
  title         = {Learning Multi-Level Hierarchies with Hindsight},
  author        = {Levy, Andrew and Konidaris, George and Platt, Robert and Saenko, Kate},
  booktitle     = {International Conference on Learning Representations},
  year          = {2019},
  eprint        = {1712.00948},
  archiveprefix = {arXiv}
}

@article{sutton1999options,
  title         = {Between {MDPs} and Semi-{MDPs}: A Framework for Temporal Abstraction in Reinforcement Learning},
  author        = {Sutton, Richard S. and Precup, Doina and Singh, Satinder},
  journal       = {Artificial Intelligence},
  volume        = {112},
  number        = {1--2},
  pages         = {181--211},
  year          = {1999},
  doi           = {10.1016/S0004-3702(99)00052-1}
}

@inproceedings{yao2014uom,
  title         = {Universal Option Models},
  author        = {Yao, Hengshuai and Szepesv{\'a}ri, Csaba and Sutton, Richard S. and Modayil, Joseph and Bhatnagar, Shalabh},
  booktitle     = {Advances in Neural Information Processing Systems},
  volume        = {27},
  pages         = {990--998},
  year          = {2014}
}

@inproceedings{kim2024openvla,
  title         = {{OpenVLA}: An Open-Source Vision-Language-Action Model},
  author        = {Kim, Moo Jin and Pertsch, Karl and Karamcheti, Siddharth and Xiao, Ted and Balakrishna, Ashwin and Nair, Suraj and Rafailov, Ethan and Foster, Ethan and Lam, Grace and Sanketi, Pannag and Vuong, Quan and Kollar, Thomas and Burchfiel, Benjamin and Tedrake, Russ and Sadigh, Dorsa and Levine, Sergey and Liang, Percy and Finn, Chelsea},
  booktitle     = {Conference on Robot Learning},
  year          = {2024},
  eprint        = {2406.09246},
  archiveprefix = {arXiv}
}

@inproceedings{black2025pi0,
  title         = {$\pi_0$: A Vision-Language-Action Flow Model for General Robot Control},
  author        = {Black, Kevin and Brown, Noah and Driess, Danny and Esmail, Adnan and Equi, Michael and Finn, Chelsea and Fusai, Niccolo and Groom, Lachy and Hausman, Karol and Ichter, Brian and Jakubczak, Szymon and Jones, Tim and Ke, Liyiming and Levine, Sergey and Li-Bell, Adrian and Mothukuri, Mohith and Nair, Suraj and Pertsch, Karl and Shi, Lucy Xiaoyang and Tanner, James and Vuong, Quan and Walling, Anna and Wang, Haohuan and Zhilinsky, Ury},
  booktitle     = {Robotics: Science and Systems},
  year          = {2025},
  doi           = {10.15607/RSS.2025.XXI.010}
}

@inproceedings{zhang2020hrac,
  title         = {Generating Adjacency-Constrained Subgoals in Hierarchical Reinforcement Learning},
  author        = {Zhang, Tianren and Guo, Shangqi and Tan, Tian and Hu, Xiaolin and Chen, Feng},
  booktitle     = {Advances in Neural Information Processing Systems},
  year          = {2020},
  eprint        = {2006.11485},
  archiveprefix = {arXiv}
}

@inproceedings{hwang2026sse,
  title         = {Strict Subgoal Execution: Reliable Long-Horizon Planning in Hierarchical Reinforcement Learning},
  author        = {Hwang, Jaebak and Lee, Sanghyeon and Kim, Jeongmo and Han, Seungyul},
  booktitle     = {International Conference on Learning Representations},
  year          = {2026},
  eprint        = {2506.21039},
  archiveprefix = {arXiv}
}

@misc{jurgenson2019subgoal,
  title         = {Sub-Goal Trees -- a Framework for Goal-Directed Trajectory Prediction and Optimization},
  author        = {Jurgenson, Tom and Groshev, Edward and Tamar, Aviv},
  year          = {2019},
  eprint        = {1906.05329},
  archiveprefix = {arXiv}
}

@misc{parascandolo2020dcmcts,
  title         = {Divide-and-Conquer {Monte} {Carlo} Tree Search for Goal-Directed Planning},
  author        = {Parascandolo, Giambattista and Buesing, Lars and Merel, Josh and Hasenclever, Leonard and Aslanides, John and Hamrick, Jessica B. and Heess, Nicolas and Neitz, Alexander and Weber, Theophane},
  year          = {2020},
  eprint        = {2004.11410},
  archiveprefix = {arXiv}
}

@inproceedings{pertsch2020gcp,
  title         = {Long-Horizon Visual Planning with Goal-Conditioned Hierarchical Predictors},
  author        = {Pertsch, Karl and Rybkin, Oleh and Ebert, Frederik and Finn, Chelsea and Jayaraman, Dinesh and Levine, Sergey},
  booktitle     = {Advances in Neural Information Processing Systems},
  year          = {2020},
  eprint        = {2006.13205},
  archiveprefix = {arXiv}
}

@inproceedings{czechowski2021ksubs,
  title         = {Subgoal Search For Complex Reasoning Tasks},
  author        = {Czechowski, Konrad and Odrzyg{\'o}{\'z}d{\'z}, Tomasz and Zbysi{\'n}ski, Marek and Zawalski, Micha{\l} and Olejnik, Krzysztof and Wu, Yuhuai and Kuci{\'n}ski, {\L}ukasz and Mi{\l}o{\'s}, Piotr},
  booktitle     = {Advances in Neural Information Processing Systems},
  year          = {2021},
  eprint        = {2108.11204},
  archiveprefix = {arXiv}
}

@misc{zhou2026idwm,
  title         = {Hierarchical World Models with Implicit Dynamics},
  author        = {Zhou, Gaoyue and Wu, Yvonne and Cui, Zichen Jeff and Ballas, Nicolas and Assran, Mido and Pinto, Lerrel and LeCun, Yann},
  year          = {2026},
  eprint        = {2610.42146},
  archivePrefix = {arXiv},
  primaryClass  = {cs.RO}
}

@misc{wang2026lmta,
  title         = {Learning Multi-Timescale Abstractions for Hierarchical Combinatorial Planning},
  author        = {Wang, Vivienne Huiling and Wang, Tinghuai and Pajarinen, Joni},
  year          = {2026},
  eprint        = {2605.17058},
  archivePrefix = {arXiv},
  primaryClass  = {cs.LG}
}

@misc{liu2026pilot,
  title         = {{PILOT}: Guiding Long-Horizon World-Model Planning for Robot Manipulation with Visual Subgoals and Policy Priors},
  author        = {Liu, Yang and Cui, Hanchen and Desingh, Karthik},
  year          = {2026},
  eprint        = {2610.14335},
  archivePrefix = {arXiv},
  primaryClass  = {cs.RO}
}

@inproceedings{ahmad2026surveyor,
  title     = {{SURVEYOR}: Certified Speculative Plan Consumption for Latent World-Model Planning},
  author    = {Ahmad, Hasaan and Jin, Gaojie and Fu, Zeyu and Yin, Lu and Huang, Tianjin},
  booktitle = {NeurIPS Workshop on World Models in Physical AI},
  year      = {2026}
}
}

\ifdefined\mainbodyonly\else
\clearpage
\appendix
\renewcommand{\thetable}{S\arabic{table}}
\renewcommand{\thefigure}{S\arabic{figure}}
\renewcommand{\theequation}{S\arabic{equation}}
\setcounter{table}{0}
\setcounter{figure}{0}
\setcounter{equation}{0}
\renewcommand{\thealgorithm}{S\arabic{algorithm}}
\setcounter{algorithm}{0}
\renewcommand{\thetheorem}{S\arabic{theorem}}
\renewcommand{\theproposition}{S\arabic{proposition}}
\renewcommand{\thelemma}{S\arabic{lemma}}
\renewcommand{\thecorollary}{S\arabic{corollary}}
\renewcommand{\thedefinition}{S\arabic{definition}}
\setcounter{theorem}{0}
\makeatletter
\setlength{\@fptop}{0pt}\setlength{\@fpsep}{14pt}\setlength{\@fpbot}{0pt plus 1fil}
\setlength{\@dblfptop}{0pt}\setlength{\@dblfpsep}{14pt}\setlength{\@dblfpbot}{0pt plus 1fil}
\makeatother
\section*{Supplementary Material}
\section{Formal Analysis}
\label{sec:proofs}

This section states the results behind the planning radius analysis of the main paper and proves them. \Cref{thm:reach} of the main paper combines Theorem~S1 and Proposition~S3, and \cref{thm:fixed} combines Proposition~S5 and Theorem~S2.

\subsection{Setting}

Consider a latent MDP on $\Zc \subset \mathbb{R}^n$ under the $\ell_1$ norm, with a null action. A planner pursues $g$ from $z$ until it enters the ball $\Ball(g)$ or exhausts its budget. The \emph{landing kernel} $\Lam_k(\cdot \mid z, g)$ is the law of the final latent of $\pik{k}$, and level $k$ plans on the MDP whose actions are subgoals and whose transitions follow $\Lam_{k-1}$.

The \emph{control distance} $D(z, g)$ is the least number of steps that can take $z$ into $\Ball(g)$. With $S_k(D) = \inf_{D(z, g) \le D} \Lam_k(\Ball(g) \mid z, g)$, the \emph{$\delta$-radius} is $r_k^\delta = \sup\{D : S_k(D) \ge 1 - \delta\}$, and $r_k = r_k^{1/2}$.

The self-model target is the coordinate-wise median $\Pstar_k = \med \Lam_{k-1}$, which minimizes the $\ell_1$ training loss, and $\Pt$ is \emph{calibrated} if it equals $\Pstar_k$. When $\Lam(\Ball(g) \mid z, g) > 1/2$, more than half of the landings lie within $\varepsilon$ of $g$ in every coordinate, so $\|\Pstar(z, g) - g\|_\infty \le \varepsilon$. A predicted landing away from $g$ thus certifies success of at most $1/2$.

The \emph{exact setting} has deterministic dynamics, $\varepsilon = 0$ and one goal candidate, so a kernel is a landing map. Level 0 runs an exact $H$-step search with a perfect $\Pz$, breaks ties toward the shortest plan, and replans after each executed prefix. Each level $k \ge 1$ is calibrated with $\st = 0$, minimizes the energy exactly over at most $M$ subgoals with all latents decodable, breaks ties toward the fewest subgoals, and makes at most $M$ calls per pursuit, so its budget is $H M^k$ steps.

\subsection{Flat Planning and Optimism}

\begin{proposition}[Flat planning]
\label{prop:flat}
On a finite reachable set, a flat planner that scores plans by latent distance, with exact search and a perfect model, executing $H$-step plans and staying put on ties, reaches $g$ if and only if its trajectory meets no $z \ne g$ from which every state within $H$ steps is no closer to $g$. It never leaves such a state.
\end{proposition}

\begin{proof}
The planner moves to the state within $H$ steps closest to $g$. At such a $z$ this is $z$ itself, so it stays forever. Otherwise the latent distance strictly decreases at every step, which on a finite set can end only at $g$. The argument does not involve prediction error.
\end{proof}

\begin{proposition}[Optimism collapses the hierarchy]
\label{prop:collapse}
Let $\st = 0$ and $\Delta_\theta = \Pt(z, g, k) - g$. Then $\zhat_M = g_M + \Delta_\theta(\zhat_{M-1}, g_M, K)$, so earlier subgoals reach the energy only through $\Delta_\theta$. If $\Delta_\theta \equiv 0$, with a decodable goal and ties toward the fewest subgoals, then $\pik{k}$ has the landing kernel of $\pik{0}$ run with the budget of level $k$, at every depth.
\end{proposition}

\begin{proof}
The first claim is the last rollout step. With $\Delta_\theta \equiv 0$, the energy equals $\|g_M - \zgoal\|_1$, which the one-subgoal plan $g_1 = \zgoal$ brings to zero, and the tie rule selects it. Each pursuit of $\pik{k}$ is therefore a sequence of runs of $\pik{k-1}$ toward the same goal. By induction these are runs of $\pik{0}$, and since Level 0 replans from its current latent alone, their concatenation is one longer run of $\pik{0}$.
\end{proof}

\subsection{Recursive Reach}

\begin{theorem}[Recursive reach and separation]
\label{thm:separation}
In the exact setting, let $r = H M^K$.
(i) $\pik{K}$ reaches $g$ from $z$ if and only if $D(z, g) \le r$.
(ii) On the path $s_0, \dots, s_r$ with unit moves, $s_i = i\, e_1$ for $i < r$ and $s_r = -e_1$, a flat planner that scores by latent distance, of horizon $H_f < r$ has radius exactly $H_f$ under any budget and never reaches $s_r$ from $s_0$, at latent distance $1$, while $\pik{K}$ reaches every goal.
(iii) With a search budget $(N, I)$ at every upper level, a decision of $\pik{K}$ costs at most $K N I M + N_0 I_0 H$ evaluations. If a budget $(N, I)$ that does not depend on $k$ realizes the exact minimization of every level, $\pik{K}$ attains the reach of (i) at this cost, while such a flat planner that reaches every goal on that path costs at least $N_f I_f\, r$, a ratio of $\Omega(r / \log r)$ at equal samples per search.
\end{theorem}

\begin{proof}
\emph{(i)} We induct on $k$. For $k = 0$, exact search finds a shortest zero-energy plan for any goal within $H$ steps. If it has length $L$, then after a prefix of $p$ steps its remainder is a zero-energy plan of length $L - p$ from the new state, so the next shortest plan has length at most $L - p$. The remaining length falls by $p$ per replan, and the pursuit arrives within $L \le H$ steps. Let the claim hold for $k - 1$ with radius $\rho = H M^{k-1}$, and let $D(z, g) \le M \rho$. Stones on a shortest path, at most $\rho$ steps apart and at most $M$ in number, are each reached by $\pik{k-1}$, and calibration makes the predicted rollout exact, so this plan has zero energy. Search returns a zero-energy plan with the fewest subgoals. After each call, its tail is a zero-energy plan from the landing, so the subgoal count falls by one per call, and the pursuit arrives within $M$ calls. Conversely, a pursuit executes at most $H M^K$ steps.

\emph{(ii)} For $i < r$ we have $\|s_i - s_r\|_1 = i + 1$. From a state with $r - i > H_f$, the goal $s_r$ is out of the planner's reach, the closest reachable state has the lowest index, and the planner descends to $s_0$ and stays. So $S(H_f + 1) = 0$, while goals within $H_f$ steps are solved by one plan. Part (i) covers $\pik{K}$.

\emph{(iii)} One upper-level search runs $N$ candidates for $I$ iterations through rollouts of $M$ steps, so it costs $N I M$ whether or not it is exact, and each of the $K$ upper levels replans at most once per decision. Under the budget condition, every level performs the exact minimization that (i) uses, so (i) holds at this cost. Every level searches $M$ codes, a problem of dimension $M d_u$ at every $k$, and Proposition~S4 states when CEM search in this dimension receives signal. By (ii), reaching every goal requires $H_f \ge r$, so each flat candidate costs at least $r$. With $K = \log_M(r / H)$, the ratio follows.
\end{proof}

Depth, and hence compute, grows logarithmically with the radius. The next result extends the radius bound to stochastic landings, where failed calls can be retried.

\begin{proposition}[Geometric radius]
\label{prop:radius}
Let $S_{k-1}(r) \ge 1 - \delta$. Let each call of a level-$k$ pursuit hand down a subgoal within control distance $r$, such that arrival lowers the distance to the goal by at least $\eta r$ or reaches the goal, and a miss raises it by at most $\xi r$, with $\eta M \in \mathbb{N}$. If $\kappa = \eta / (\eta + \xi) > \delta$ and the budget is $n = \lceil 2 \kappa M / (\kappa - \delta) \rceil$ calls, then $S_k(\eta M r) \ge 1 - e^{-\kappa (\kappa - \delta) M}$. If this bound is at least $1 - \delta$ at every level, then $r_K^\delta \ge r_0^\delta (\eta M)^K$.
\end{proposition}

\begin{proof}
Let $f$ count the misses. If the pursuit has not arrived after $n$ calls, summing the changes in control distance gives $0 < \eta M r - \eta r (n - f) + \xi r f$, that is, $f > \kappa (n - M) \ge n (\kappa + \delta) / 2$. Each call misses with conditional probability at most $\delta$, so the Azuma--Hoeffding inequality bounds this event by $e^{-n (\kappa - \delta)^2 / 2} \le e^{-\kappa (\kappa - \delta) M}$. Induction over levels gives the radius.
\end{proof}

\subsection{Search Transition}

\begin{proposition}[Search transition]
\label{prop:transition}
Let a plan $u = u_{1:M}$ have dimension $d = M d_u$. Let the energy increase strictly with $\|u - u^\star\|_2$ on the ball of radius $\rho$ and be constant outside it. Let $u = u^\star + e + s$, with prior mean error on the scale of the prior width, $e \sim \mathcal{N}(0, \sq^2 I)$, and independent sample noise $s \sim \mathcal{N}(0, \sq^2 I)$, and let $d_c = \rho^2 / (2 \sq^2)$ and $t \in (0, 1)$.
(i) If $d \le d_c / (1 + t)$, all $N$ samples lie inside the ball with probability at least $1 - N e^{-d t^2 / 8}$, and the elite refit then moves toward $u^\star$ in expectation.
(ii) If $d \ge d_c / (1 - t)$, all $N$ samples lie outside with probability at least $1 - N e^{-d t^2 / 4}$, and the refit then does not move toward $u^\star$ in expectation.
\end{proposition}

\begin{proof}
Marginally, $\|u - u^\star\|_2^2 = 2 \sq^2 \chi^2_{d}$, and the Chernoff bounds $\Pr[\chi^2_d \ge d(1 + t)] \le e^{-d t^2 / 8}$ and $\Pr[\chi^2_d \le d(1 - t)] \le e^{-d t^2 / 4}$ with a union bound give both probabilities. In (i), elites are the samples closest to $u^\star$, so given $e$, their noise $s$ is restricted to a ball centered at $-e$. Along the direction of $e$ this is an interval centered on the negative side, and the even Gaussian density gives a negative mean, so the elite mean moves toward $u^\star$. In (ii), all energies are equal, so the elites are a subset chosen independently of the noise, and their mean is the mean of $s$ conditioned on lying outside that ball. The excluded ball is centered on the negative side along the direction of $e$, so the remaining density has a nonnegative mean along it, and the refit does not move toward $u^\star$.
\end{proof}

\subsection{Grounded Self-Improvement}

\begin{proposition}[Selection bias and correction]
\label{prop:drift}
Let $N$ candidates have true energies of spread $\Delta$ and predictions $\hat E_i = E_i - b - \sigma \zeta_i$, with independent zero-mean 1-sub-Gaussian $\zeta_i$.
(i) Selecting the minimum of $\hat E_i$ adds a bias $b_{\mathrm{s}} = \sigma \Ex[\zeta_{i^\star}] \in [0, \sigma \sqrt{2 \ln N}]$, with $b_{\mathrm{s}} \ge \sigma \Ex[\max_i \zeta_i] - \Delta$.
(ii) If imagined labels pass a fraction $c \in (0, 1]$ of this bias to the next model, the bias $b_n$ after $n$ rounds never decreases.
(iii) With $\pik{k-1}$ frozen, target $f^\star \in \Fc$ and each retrained $f_t \in \Fc$ consistent with all earlier executed outcomes, at most $\dimE(\Fc, \alpha)$ retraining steps select a query with error above $\alpha$, under any selection rule.
\end{proposition}

\begin{proof}
\emph{(i)} The event $\{i^\star = i\}$ is nondecreasing in $\zeta_i$, so $\Ex[\zeta_{i^\star}] \ge 0$ by association. Also $\zeta_{i^\star} \le \max_i \zeta_i$, and $\Ex \max_i \zeta_i \le \sqrt{2 \ln N}$ by the standard moment bound. Optimality of $i^\star$ against the maximizer of $\zeta_i$ gives the lower bound. \emph{(ii)} This follows from $b_{n+1} = b_n + c\, b_{\mathrm{s},n}$ and $b_{\mathrm{s},n} \ge 0$. \emph{(iii)} Since $f_t$ and $f^\star$ agree on earlier queries, a query with error above $\alpha$ is $\alpha$-independent of its predecessors~\cite{russo2013eluder}, and such a sequence has length at most $\dimE(\Fc, \alpha)$. With a fitting error of $\beta$, the same argument gives $(\beta / \alpha^2 + 1) \dimE(\Fc, \alpha)$. The count relies on observing the true landing at each selected query, which imagined labels never provide.
\end{proof}

\begin{theorem}[Grounded fixed point]
\label{thm:grounded}
In the exact setting, let every $(z, g)$ receive an executed tuple in each round, and let a round return at every input a minimizer of the weighted $\ell_1$ loss over the labels of that input, ties broken by a fixed rule. Let $\Phi$ map self-models $(P_k)_{k \le K}$ to their values retrained on the round's executed outcomes and their hindsight relabels.
(i) $\Phi$ is well defined from every initialization, has the unique fixed point $(\Pstar_k)_{k \le K}$, and $\Phi^K$ sends every initialization to it.
(ii) From $P_k = g$, with one level unlocked per round, the planner after round $t \le K$ reaches every goal with $D(z, g) \le H M^t$.
(iii) Under labels imagined by the self-model itself, every idempotent $P_k$, with $P_k(z, P_k(z, g)) = P_k(z, g)$, is a fixed point, including $P_k = g$.
\end{theorem}

\begin{proof}
\emph{(i)} The labels of $P_k$ at an input $(z, g)$ are the executed landing $\Lam_{k-1}(z, g)$ and, if a pursuit from $z$ landed at $g$, the hindsight label $g$. Before the levels below are calibrated these may disagree, and the weighted $\ell_1$ minimizer with its tie rule still returns one value, so $\Phi$ is well defined. Both labels are determined by the landing map of level $k{-}1$, which depends only on $P_{1:k-1}$, so $\Phi(P)_k$ depends only on $P_{1:k-1}$ and the update is triangular. With $P_{1:k-1}$ calibrated, the labels agree: a landing $g$ of $\pik{k-1}$ from $z$ lies within $H M^{k-1}$ steps of $z$, so $\Lam_{k-1}(z, g) = g$ by Theorem~S1(i). Every label at $(z, g)$ then equals $\Lam_{k-1}(z, g) = \Pstar_k(z, g)$, and so does the minimizer. Induction over $k$ gives $\Phi^t(P)_k = \Pstar_k$ for $k \le t$, regardless of the earlier $P_k$. A fixed point $P$ satisfies $P = \Phi^K(P) = \Pstar$, which gives uniqueness. \emph{(ii)} After round $t$, levels up to $t$ are calibrated, and Theorem~S1 gives the reach. The newly unlocked level is still at $P = g$, so by Proposition~S2 it delegates to level $t$. \emph{(iii)} The executed tuple has label $P_k(z, g)$, and the hindsight tuple has the same label by idempotence, so the labels agree and the fit returns $P_k$.
\end{proof}

\section{Implementation}
\label{sec:impl}

\subsection{Modules}

\paragraph{Observations and Latents.}
$E$ is the V-JEPA 2 encoder applied per camera view, and tokens are concatenated across views to give $z_t$. Every $\ell_1$ distance is averaged over tokens and dimensions.

\paragraph{World Model.}
$\Pz$ follows V-JEPA 2-AC~\cite{assran2025vjepa2} and also receives the end-effector state $s_t$ and a short frame context. With two views, a learned view embedding is added to the tokens of each view after the input projection, the views share the spatial grid and the time index of the rotary position encoding, and the block-causal mask lets the tokens of a step attend to both views at that step and at earlier ones, so one set of weights predicts the next latent of both views. During rollout, $s$ is propagated with the planned actions. The training loss combines teacher-forcing and rollout steps against stop-gradient target latents. In simulation, $\Pz$ trains on the demonstrations of each benchmark. On the robot, it is fine-tuned from the checkpoint trained on DROID~\cite{khazatsky2024droid}, which has one external view: the view embeddings start at zero, so the external view enters as in the checkpoint, and the state and action embeddings carry over, since the action is the same seven-dimensional delta end-effector pose and gripper command. On LeWM~\cite{maes2026lewm}, $z_t$ is a single token, $E$ and $\Pz$ train jointly from pixels with the SIGReg regularizer, $\Pt$ copies the LeWM predictor, and $\Dp$ is an MLP.

\paragraph{Self-Model.}
$\Pt$ is initialized from the weights of $\Pz$. The action embedding is replaced by a cross-attention embedding of the tokens of $g$, the state embedding is removed, and a level embedding for $k \in \{1, \dots, K\}$ is added. Each output head is an MLP with its own initialization and a per-tuple bootstrap mask. The uncertainty $\st$ is the standard deviation across heads, averaged over tokens, and the coordinate-wise median of the heads is the prediction. Stage 1 samples a demonstration, a start $t$ and an offset $h \sim \mathcal{U}[1, h_k]$ with $h_k = h_1 M^{k-1}$, and trains on the tuple $(z_t, z_{t+h}, z_{t+h}, k)$ for every $k$.

\paragraph{Subgoal Code.}
$\Ep$ pools tokens by attention into $u \in \mathbb{R}^{d_u}$, and $\Dp$ reconstructs the tokens by cross-attention from learned queries, under an $\ell_1$ loss. Codes are standardized with per-dimension statistics of the demonstration codes, so that the unit Gaussian $\mathcal{N}(0, I)$ matches the code marginal, and samples are clipped to per-dimension quantiles~\cite{zhang2026hwm}. Each round fine-tunes on demonstration and reached latents together, so decoded subgoals cover both visited states and the goal region.

\paragraph{Prior.}
$\qp(u \mid z, \zgoal, k)$ is a diagonal Gaussian from an MLP over pooled tokens and a level embedding. Replay stores latents, and codes are recomputed with the current $\Ep$ at each fit. Its targets are the codes of hindsight landings and of first subgoals whose executed landing lies near the aim.

\paragraph{Goal Head.}
$\Gw$ is a small transformer over text features, $z_0$ and noise $w_j \sim \mathcal{N}(0, I)$, following the language-to-latent prediction of LAGO~\cite{barbeau2026lago}. It runs once per episode.

\paragraph{Visualization Decoder.}
Decoded subgoals in the figures come from a ViT decoder trained on encoder latents with a pixel loss, as in HWM~\cite{zhang2026hwm}. Planning does not use it.

\Cref{tab:stages} gives the training status of each module, and \cref{tab:hyper} lists the hyperparameters.

\begin{table}[t]
  \centering
  \caption{\textbf{Training stages.} Modules and how each stage treats them. Stage 1 uses demonstrations, and Stage 2 uses executed outcomes. The world model stays fixed in Stage 2, so the upper levels account for all improvement.}
  \label{tab:stages}
  \small
  \setlength{\tabcolsep}{4pt}
  \begin{tabular}{lll}
    \toprule
    Module & Stage 1 & Stage 2 \\
    \midrule
    Encoder $E$ & Pretrained & Frozen \\
    World model $\Pz$ & Trained & Frozen \\
    Text encoder & Pretrained & Frozen \\
    Goal head $\Gw$ & Trained & Frozen \\
    \midrule
    Self-model $\Pt$ & Pretrained on hindsight pairs & Trained \\
    Code $(\Ep, \Dp)$ & Pretrained & Fine-tuned \\
    Prior $\qp$ & Pretrained & Trained \\
    \bottomrule
  \end{tabular}
\end{table}

\begin{table}[t]
  \centering
  \caption{\textbf{Hyperparameters.} Values used across benchmarks.}
  \label{tab:hyper}
  \footnotesize
  \begin{tabular}{ll}
    \toprule
    Hyperparameter & Value \\
    \midrule
    \multicolumn{2}{l}{\emph{World model}} \\
    Context, teacher-forcing, rollout steps $L$ & 2, 8, 2 \\
    Learning rate ($10^{-4}$), batch size, steps (k) & 2, 256, 60 \\
    \midrule
    \multicolumn{2}{l}{\emph{Self-model}} \\
    Output heads, head depth, bootstrap rate & 5, 2, 0.8 \\
    Learning rate ($10^{-4}$), Stage 1 steps, steps per round (k) & 3, 100, 10 \\
    Stage 1 offset $h_1$ & 24 \\
    \midrule
    \multicolumn{2}{l}{\emph{Subgoal code and prior}} \\
    Code dimension $d_u$ & 8 \\
    Code Stage 1 steps, steps per round (k) & 50, 5 \\
    Clip quantiles (\%) & 0.5, 99.5 \\
    Prior learning rate ($10^{-4}$), steps per round (k) & 3, 5 \\
    \midrule
    \multicolumn{2}{l}{\emph{Hierarchy}} \\
    Depth $K$, subgoals per plan $M$ & 3, 3 \\
    Uncertainty weight $\lambda$, delegation margin $\nu$ & 0.5, 0.02 \\
    Arrival $\epsilon_k$, Level-0 budget $T_0$ ($T_k = T_0 M^k$) & 0.05, 16 \\
    \midrule
    \multicolumn{2}{l}{\emph{Search}} \\
    Upper-level CEM $(N, I)$ & 64, 3 \\
    Elites $N_e$ & 8 \\
    Level 0 CEM $(N_0, I_0)$ & 48, 3 \\
    Action horizon $H$, executed prefix $p$ & 8, 4 \\
    \midrule
    \multicolumn{2}{l}{\emph{Self-improvement}} \\
    Rounds $R$, episodes per round (simulation) & 5, 2{,}000 \\
    Unlock threshold $\tau$, target offset $D_1$ & 0.6, 48 \\
    Exploration probability & 0.1 \\
    Hindsight fraction, recency weight $\gamma$ & 0.5, 0.7 \\
    \midrule
    \multicolumn{2}{l}{\emph{Goal head}} \\
    Goal candidates $J$ & 4 \\
    \bottomrule
  \end{tabular}
\end{table}

\subsection{Recursive Planner}

\Cref{alg:pursue,alg:search} specify the planner. A decision is one Level 0 replan. A level holds its subgoal for the whole call to the level below.

\begin{algorithm}[t]
  \caption{Recursive pursuit}
  \label{alg:pursue}
  \small
  \begin{algorithmic}[1]
    \Function{Pursue}{$j, g$}
      \State $z_s \gets E(o)$, $b \gets 0$
      \While{$\|E(o) - g\|_1 > \epsilon_j$, $b < T_j$, episode active}
        \If{$j = 0$}
          \State $a^\star \gets$ CEM on $\|\zhat_{t+H} - g\|_1$ with $(N_0, I_0)$
          \State execute the first $p$ actions, $b \gets b + p$
        \Else
          \State $(u^\star, e^\star) \gets$ \Call{SearchCodes}{$E(o), g, j$}
          \State $e_{\mathrm{d}} \gets \|\Pt(E(o), g, j) - g\|_1 + \lambda \st(E(o), g, j)$
          \State $g' \gets g$ \textbf{if} $e_{\mathrm{d}} \le e^\star + \nu$ \textbf{else} $\Dp(u^\star_1)$
          \State $b \gets b\, +$ steps of \Call{Pursue}{$j{-}1, g'$}
        \EndIf
      \EndWhile
      \State in a round, add $(z_s, g, E(o), j{+}1)$ to $\replay$
      \State \Return $E(o)$
    \EndFunction
  \end{algorithmic}
\end{algorithm}

\begin{algorithm}[t]
  \caption{Code search at level $k \ge 1$}
  \label{alg:search}
  \small
  \begin{algorithmic}[1]
    \Function{SearchCodes}{$z, g, k$}
      \State draw $N/2$ plans by ancestral sampling, $u_m \sim \qp(\cdot \mid \zhat_{m-1}, g, k)$
      \State draw $N/2$ plans with $u_m \sim \mathcal{N}(0, I)$
      \For{$i = 1, \dots, I$}
        \State score all plans with the energy, rolling out $\Pt$
        \State fit a diagonal $\mathcal{N}(\mu, \Sigma)$ to the $N_e$ best plans
        \State \textbf{if} $i < I$, draw $N$ plans from it
      \EndFor
      \State \Return $\mu$ and its energy
    \EndFunction
  \end{algorithmic}
\end{algorithm}

A level hands the goal down when its self-model predicts no better landing through intermediate subgoals. The pretrained model satisfies this rule at every state, so round 0 equals $\pik{0}$. The top-level call is $\textsc{Pursue}(K, \cdot)$, and it takes the nearest of the $J$ candidate goals. The episode ends on benchmark success or at the benchmark horizon. Level 0 initializes its mean from the shifted tail of the previous solution.

\subsection{Self-Improvement}

\Cref{alg:loop} lists the loop of \cref{sec:loop} of the main paper.

\begin{algorithm}[t]
  \caption{Grounded self-improvement}
  \label{alg:loop}
  \small
  \begin{algorithmic}[1]
    \Require frozen $E, \Pz, \Gw$; demonstrations $\mathcal{D}$
    \State Pretrain $\Pt$ on hindsight pairs from $\mathcal{D}$; $\Ep, \Dp, \qp$ on $\mathcal{D}$
    \State $\replay \gets \emptyset$, $\bar k \gets 1$
    \For{round $= 1, \dots, R$}
      \State \textbf{Execute:} run the planner of depth $\bar k$; add each $(z, g, \zreach, k)$ to $\replay$
      \State \textbf{Hindsight:} add $(z, \zreach, \zreach, k)$ to $\replay$
      \State \textbf{Retrain:} $\Pt$ with the self-model loss; $\Ep, \Dp, \qp$ on $\replay$
      \State \textbf{Unlock:} if $S_{\bar k}(D_{\bar k}) \ge \tau$ and $\bar k < K$, set $\bar k \gets \bar k + 1$
    \EndFor
  \end{algorithmic}
\end{algorithm}

\paragraph{Execute.}
Each episode samples a task uniformly and an initial state from the training pool. With a small probability, an upper level takes $\Dp(u^\star_1)$ instead of delegating, which yields landing data for intermediate subgoals from round 1 onward. Every pursuit at every level, including the top-level one, adds a tuple. Top-level tuples carry index $K{+}1$, so a newly unlocked level starts with data. Tuples store latents, and pursuits cut off by the episode end are discarded.

\paragraph{Hindsight.}
Each tuple adds $(z_s, \zreach, \zreach, k)$, and sub-segments of the pursuit add further pairs.

\paragraph{Retrain.}
Replay starts empty after Stage 1, and all modules warm-start from the previous round. Mini-batches are balanced over unlocked levels, hold a fixed fraction of hindsight tuples, and weight a tuple from $n$ rounds ago by $\gamma^n$.

\paragraph{Unlock.}
Level $k$ has a target offset $D_k = D_1 M^{k-1}$ steps. Level $k{+}1$ opens once $\pik{k}$ succeeds at $D_k$ with rate $\tau$ on offset goals from the training pool, up to depth $K$.

\section{Protocol}
\label{sec:protocol}

\subsection{Benchmarks and Data}

\paragraph{LIBERO.}
We use the ten tasks of each suite~\cite{liu2023libero} with the published demonstrations, language instructions and fixed evaluation initial states, at the benchmark horizon. Actions are the benchmark's delta end-effector pose and gripper command. Success is the BDDL goal predicate. Self-improvement draws initial states from the task's placement distribution and rejects any close to an evaluation state.

\paragraph{RoboCasa.}
We use the 24 atomic tasks of RoboCasa~\cite{nasiriany2024robocasa}, grouped by skill as in the benchmark, and its demonstration set, with the base held fixed. Evaluation follows the benchmark over its fixed evaluation scenes, disjoint from the scenes used for self-improvement.

\paragraph{Goal Images.}
For a LIBERO evaluation state, the goal image is the final frame of the training demonstration of that task whose initial object layout is nearest in simulator state. RoboCasa uses the same rule within the task.

\paragraph{Interaction Budget.}
The budget is $R$ rounds per suite with about 2{,}000 episodes per round, shared equally across the tasks of the suite. Models are trained separately per suite on LIBERO and per task group on RoboCasa.

\subsection{Metrics}

\paragraph{Success Rate.}
The rate per task is the mean over evaluation episodes. A suite score is the unweighted mean over tasks and seeds, where a seed covers Stage 1, all rounds and evaluation.

\paragraph{Planning Radius.}
A query restores the simulator state of frame $t$ of a demonstration and takes frame $t{+}D$ as the goal image. Offsets $D$ are in control steps, and the horizon of a query is $2D$ steps. A query succeeds when task-relevant object poses and the gripper state lie within fixed position and angle tolerances of those of the goal frame. $\rfifty$ is the largest $D$ with success at least 50\%, linearly interpolated between offsets, up to the longest demonstration.

\paragraph{Aim--Arrive Gap.}
This is the mean of $\|\zhat' - \zreach\|_1$ over all upper-level pursuits in evaluation, with $\zhat'$ the first predicted landing of the selected plan at hand-off.

\paragraph{Compute.}
One model evaluation is one candidate advanced one step by $\Pz$ or $\Pt$, regardless of the head count. A rollout of length $L$ costs $L$. Evaluations per decision are the episode total over the episode's Level 0 replans, and \cref{tab:depth} also reports the total per episode. Time per decision covers encoding, $\Dp$, $\qp$ and graph search, on one GPU. The compute curves vary $(N, N_0)$ for RS-JEPA and the corresponding budget for each baseline.

\paragraph{Search Transition.}
$\rho$ is the largest code-space radius around the elite mean within which the rank correlation of energy and executed landing distance stays above a fixed threshold, measured once at the selected $d_u$ and held fixed across the sweep. $\sq$ is the mean prior width over evaluation states.

\subsection{Baselines}

\paragraph{Shared Components.}
All planners use the same $E$, and all goal-conditioned planners use the same $\Gw$. World-model planners share $\Pz$ at their lowest level. Each round's Dyna pass fine-tunes $\Pz$ on the method's own executed episodes and then retrains its planner modules.

\paragraph{Hyperparameter Selection.}
V-JEPA 2-AC selects its horizon and sample count on validation initial states. Validation initial states are drawn from each task's placement distribution and are disjoint from the fixed evaluation states. RS-JEPA selects $d_u$, $M$, $K$, $\tau$, $\lambda$, $\nu$ and $D_1$ once on validation initial states of LIBERO-Goal and keeps them for every suite (\cref{tab:hyper}). Each baseline tunes its own hyperparameters on the same validation initial states with a sweep of the same size.

\paragraph{Adaptation.}
\Cref{tab:ports} gives the adaptation of each method. V-JEPA 2-AC + RC-aux adds the budget-conditioned reachability loss of RC-aux~\cite{li2026rcaux} to $\Pz$, plans with the same flat CEM on the reachability-aware energy, and selects its horizon on validation initial states. The oracle reference gives Level 0 the latents of the demonstration whose initial layout is nearest, one every $H$ steps, with no upper level.

\paragraph{Published Settings.}
SAGE evaluates on two LIBERO scenes built for its study, with five and three tasks, where it reaches 48.7\% and 58.0\% of full episodes from goal images on LeWM with additional oracle data~\cite{cheng2026sage}; the standard suites and language protocol used here form a different setting, and no baseline reports results on them.

\begin{table}[t]
  \centering
  \caption{\textbf{Baseline adaptation.} Published backbone, adaptation to the shared encoder, and use of the interaction budget.}
  \label{tab:ports}
  \footnotesize
  \setlength{\tabcolsep}{3pt}
  \begin{tabular}{p{1.25cm}p{1.15cm}p{2.85cm}p{2.15cm}}
    \toprule
    Method & Published & Adaptation & Budget \\
    \midrule
    V-JEPA 2-AC & Same & Horizon and samples selected on validation & Dyna on $\Pz$ \\
    HWM & Same & Action encoder and long-horizon predictor on $E$ & Dyna on both predictors \\
    H-JEPA & Own & Level 0 shares $E, \Pz$; upper encoders are heads over $E$ tokens & Dyna on all levels \\
    + RC-aux & Same & Reachability loss on $\Pz$, flat CEM & Dyna on $\Pz$ \\
    SAGE & LeWM & Subgoal and action generators on $z_t$ & Dyna on $\Pz$; generators retrained with hindsight \\
    Metro-WM & LeWM, PLDM & Graph over demonstration latents, with $\Pz$ and CEM below & Dyna on $\Pz$; executed frames join the graph \\
    RP1 & LeWM, PLDM & Critic and operator through $\Pz$, retrained per horizon & Dyna, then retraining \\
    FF-JEPA & JEPA & One subgoal predictor per task, goal-free & None (reference) \\
    \bottomrule
  \end{tabular}
\end{table}

\subsection{Ablation Variants}

\emph{Self-model $\to$ world model:} the upper level is an HWM predictor over macro-action codes, trained on its own executed rounds under the same budget, with the same CEM budget and a prior over macro-action codes, so the row changes the prediction target and keeps the budget. \emph{Executed $\to$ imagined outcomes:} Execute runs the same hierarchy inside $\Pz$, and each landing is the $\Pz$ rollout under the selected actions; hindsight, retraining and episode counts stay the same, unlock uses imagined success, and $\rfifty$ is measured in the simulator after each round. \emph{w/o self-improvement:} the pretrained modules at depth 1, with Level 0 at the budget of the flat planner. \emph{w/o hindsight:} hindsight tuples are omitted, and Stage 1 stays. \emph{Self-model $\to$ prior mean:} the upper level hands down the decoded mean of $\qp(\cdot \mid z, \zgoal, k)$ as its subgoal, without self-model scoring or search; the rounds stay the same, and $\qp$ trains on hindsight codes and on first subgoals that the level below reached. \emph{Learned prior $\to$ Gaussian:} all samples come from $\mathcal{N}(0, I)$, with $(N, I)$ raised until success matches the full model, and the row reports the resulting compute. \emph{Frozen $\Pz$ $\to$ Dyna on $\Pz$:} each round also fine-tunes $\Pz$ on executed episodes, as the baselines do. \emph{w/o uncertainty:} $\lambda = 0$. \emph{Shared $\to$ per-level:} one network per level with the same architecture and pretraining.

\subsection{Real Robot Protocol}

\paragraph{Hardware.}
The arm is a Seeed reBot Arm B601-RS with a parallel gripper, commanded in delta end-effector pose through inverse kinematics. An Intel RealSense D435, using its color stream only, stays fixed, and an e-con Systems See3CAM\_24CUG sits on the wrist. $E$ encodes both views with their tokens concatenated.

\paragraph{Demonstrations.}
Leader--follower teleoperation provides 30 demonstrations per task, 180 in total, none from the evaluation configurations. One set of models trains on the demonstrations of all six tasks and serves every task.

\paragraph{Stage 1.}
$\Pz$ starts from the V-JEPA 2-AC checkpoint trained on DROID and is fine-tuned on these demonstrations for 20k steps with the encoder frozen, taking the wrist view through the view embedding of \cref{sec:impl}. $\Pt$, $(\Ep, \Dp)$ and $\qp$ then pretrain on the same demonstrations as in simulation, and every planner starts from this $\Pz$. The simulated benchmarks and the real robot share no data or weights beyond the pretrained encoder.

\paragraph{Tasks.}
Six tasks, each with 4 start--goal configurations and 5 trials per configuration, 20 trials per task and 120 per method. \emph{Stack cups:} place one paper cup inside the other. \emph{Blocked drawer:} a cup stands in front of a closed drawer, and the goal image shows the drawer open; success means the drawer reaches the goal extension. \emph{Object in drawer:} a cube lies inside a closed transparent drawer, and success means the cube rests on the table at the goal position. \emph{Clean table:} a pepper, a tomato and a potato go onto a plate, and success requires all three. \emph{Rings on peg:} a round ring and then a square ring go onto a peg, and success requires both in this order. \emph{Peg insertion:} the gripper holds a peg and inserts it into a hole; after the 3 rounds the planner halves the Level-0 action limit, every method practices for two further rounds on this task, and success is then measured. The operator scores each trial against marked goal positions with the tolerances of \cref{tab:protocol}, and an abort at a workspace limit counts as failure.

\paragraph{Self-Improvement and Evaluation.}
RS-JEPA runs 3 self-improvement rounds on the robot, each of 120 episodes shared equally across the six tasks, with goals from 30 training configurations disjoint from the evaluation set and with operator resets. The rounds train $\Pt$, $(\Ep, \Dp)$ and $\qp$ on the executed outcomes, and $\Pz$ stays frozen. V-JEPA 2-AC, HWM, H-JEPA and Metro-WM run Dyna passes on their world models with the same number of robot episodes of their own, their upper-level modules train on the same demonstrations, and the Metro-WM graph is built over them. Evaluation trials interleave methods in randomized order per configuration, starting from a marked object placement.

\subsection{Protocol Values}

\Cref{tab:protocol} lists the values used by this section.

\begin{table}[t]
  \centering
  \caption{\textbf{Protocol values.} Rendering, evaluation, metrics, real-robot settings and compute.}
  \label{tab:protocol}
  \footnotesize
  \begin{tabular}{ll}
    \toprule
    Item & Value \\
    \midrule
    \multicolumn{2}{l}{\emph{Observations}} \\
    Frame size (px), simulated views & 256, 2 \\
    Encoder, tokens per view, token dimension & ViT-g/16, 256, 1408 \\
    Text encoder & T5-Base \\
    \midrule
    \multicolumn{2}{l}{\emph{Evaluation}} \\
    LIBERO episodes per task & 50 \\
    Horizon: Spatial, Object, Goal, Long & 220, 280, 300, 520 \\
    RoboCasa tasks, trials, scenes & 24, 50, 5 \\
    Seeds & 3 \\
    Flat planner horizon, samples & 20, 256 \\
    \midrule
    \multicolumn{2}{l}{\emph{Planning radius}} \\
    Offset step, queries per offset & 10, 50 \\
    Position (cm), angle ($^\circ$) tolerance & 2, 10 \\
    Longest demonstration & 480 \\
    \midrule
    \multicolumn{2}{l}{\emph{Real robot}} \\
    Degrees of freedom, control rate (Hz) & 6, 10 \\
    Demonstrations per task, total & 30, 180 \\
    $\Pz$ fine-tuning steps (k) & 20 \\
    Rounds, episodes per round, total & 3, 120, 360 \\
    Training configurations, steps per trial & 30, 300 \\
    Pick-and-place, drawer tolerance (cm) & 2, 1 \\
    \midrule
    \multicolumn{2}{l}{\emph{Compute}} \\
    Stage 1, round, robot round (GPU-h) & 96, 40, 12 \\
    Dyna steps per round (k) & 5 \\
    \bottomrule
  \end{tabular}
\end{table}

\section{Comparison with Closest Work}
\label{sec:closest}

\Cref{tab:related} compares RS-JEPA with the closest latent world-model planners, outcome models and hindsight-based hierarchical reinforcement learning methods along the axes of \cref{sec:related} of the main paper: whether an upper level predicts the outcome of the level below, whether that model trains on executed outcomes and keeps improving after deployment, and whether the hierarchy recurses beyond two levels.

\begin{table}[t]
  \centering
  \caption{\textbf{Comparison with closest work.} Columns: subgoal hierarchy (Hier.), lowest level plans with a latent world model (WM), upper level predicts the lower level's outcome (Outcome), trained on executed outcomes (Exec.), repeated improvement after deployment (Repeat), and recursive depth beyond two levels (Depth).}
  \label{tab:related}
  \small
  \setlength{\tabcolsep}{2.8pt}
  \begin{tabular}{lcccccc}
    \toprule
    Method & Hier. & WM & Outcome & Exec. & Repeat & Depth \\
    \midrule
    HWM~\cite{zhang2026hwm} & \cmark & \cmark & \xmark & \xmark & \xmark & \xmark \\
    H-JEPA~\cite{zhang2026hjepa} & \cmark & \cmark & \xmark & \xmark & \xmark & \cmark \\
    Hi-LeWM~\cite{caselli2026hilewm} & \cmark & \cmark & \xmark & \xmark & \xmark & \xmark \\
    ID-WM~\cite{zhou2026idwm} & \cmark & \cmark & \xmark & \xmark & \xmark & \xmark \\
    Metro-WM~\cite{lee2026metro} & \cmark & \cmark & \xmark & \xmark & \xmark & \xmark \\
    SAGE~\cite{cheng2026sage} & \cmark & \cmark & \xmark & \xmark & \xmark & \xmark \\
    PILOT~\cite{liu2026pilot} & \cmark & \cmark & \xmark & \xmark & \xmark & \xmark \\
    SURVEYOR~\cite{ahmad2026surveyor} & \cmark & \cmark & \xmark & \xmark & \xmark & \xmark \\
    RC-aux~\cite{li2026rcaux} & \xmark & \cmark & \xmark & \xmark & \xmark & \xmark \\
    RP1~\cite{sommer2026rp1} & \xmark & \cmark & \xmark & \cmark & \xmark & \xmark \\
    SkiMo~\cite{shi2022skimo} & \cmark & \xmark & \cmark & \cmark & \cmark & \xmark \\
    LMTA~\cite{wang2026lmta} & \cmark & \xmark & \cmark & \cmark & \cmark & \xmark \\
    LEAP~\cite{nasiriany2019leap} & \cmark & \xmark & \cmark & \cmark & \xmark & \xmark \\
    HAC~\cite{levy2019hac} & \cmark & \xmark & \xmark & \cmark & \cmark & \cmark \\
    HRAC~\cite{zhang2020hrac} & \cmark & \xmark & \xmark & \cmark & \cmark & \xmark \\
    SSE~\cite{hwang2026sse} & \cmark & \xmark & \xmark & \cmark & \cmark & \xmark \\
    \midrule
    RS-JEPA (Ours) & \cmark & \cmark & \cmark & \cmark & \cmark & \cmark \\
    \bottomrule
  \end{tabular}
\end{table}

\section{Additional Results}
\label{sec:more}

\Cref{tab:depth} reports reach and compute against depth, with the two predictors counted separately. The radius grows geometrically while compute grows at most linearly. \Cref{tab:offline} lists results with no interaction, with goal images and for reference methods, \cref{tab:split} the LIBERO-Long split by task length, and \cref{tab:ablimpl} the remaining ablations. \Cref{fig:dsweep} tests the search transition of Proposition S4 and tracks the prior and the energies over rounds, \cref{fig:search} shows one search per planner, \cref{fig:qual} shows LIBERO episodes, and \cref{fig:qualdetour,fig:quallong} show real-robot episodes against V-JEPA 2-AC.

\Cref{tab:backbone} compares two JEPA world models: V-JEPA 2-AC, with a frozen pretrained encoder, and LeWM, which trains its encoder end to end from pixels and has a far more compact latent. The method uses only $E$ and $\Pz$, and on each backbone the same order holds: flat planning, a hierarchical world model, then the self-model. The self-model adds the most on LIBERO-Long on both backbones, and the order holds on every suite.

\begin{table}[t]
  \centering
  \caption{\textbf{Reach and compute against depth} on LIBERO-Long. Radius grows geometrically while evaluations per decision grow at most linearly, and upper levels replan rarely. $K = 0$ is the small Level-0 CEM of RS-JEPA ($H = 8$, $N_0 = 48$); the tuned flat planner in the main paper uses $H = 20$ and $N = 256$.}
  \label{tab:depth}
  \footnotesize
  \setlength{\tabcolsep}{3pt}
  \begin{tabular}{lccccc}
    \toprule
    $K$ & $\rfifty$ $\uparrow$ & $\Pz$ evals $\downarrow$ & $\Pt$ evals $\downarrow$ & \shortstack{Upper replans\\per episode} & \shortstack{Evals per\\episode (k) $\downarrow$} \\
    \midrule
    0 & 27 & 1152 & \textbf{0} & 0 & \textbf{104} \\
    1 & 61 & 1152 & \underline{196} & 23 & \underline{125} \\
    2 & \underline{109} & 1152 & 249 & 30 & 133 \\
    3 & \textbf{179} & 1152 & 280 & 32 & 127 \\
    \bottomrule
  \end{tabular}
\end{table}

\begin{figure*}[tp]
  \centering
  \includegraphics[width=\textwidth]{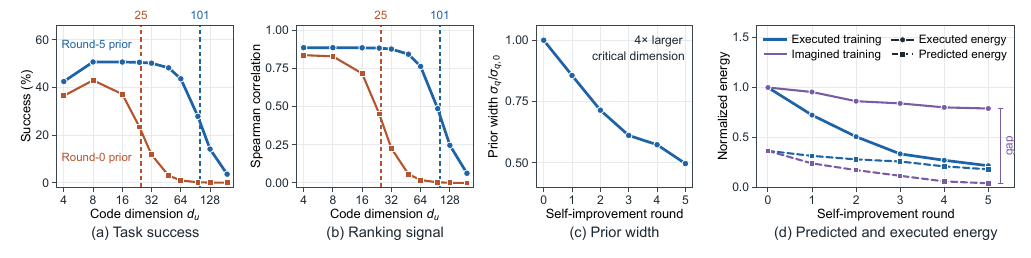}
  \caption{\textbf{Search transition and self-improvement} on LIBERO-Long.
  (a, b) A controlled prior comparison holds the round-5 self-model and CEM budget fixed and swaps the round-0 and round-5 priors. Success (a) and the Spearman correlation between predicted energy and executed landing distance (b) decrease near the predicted code dimension $d_u^*=\rho^2/(2M\sq^2)$ (dashed lines), with $M=3$ and a shared signal radius $\rho$. The round-5 point at $d_u=8$ is the LIBERO-Long result of RS-JEPA in the main paper.
  (c) Executed training narrows the prior, moving the threshold to larger dimensions.
  (d) Energies share the round-0 executed-energy normalization. Executed labels close the prediction--execution gap; imagined labels reduce predicted energy while executed energy remains high. The bracket marks this gap for imagined training at round 5.}
  \label{fig:dsweep}
\end{figure*}

\begin{figure*}[tp]
  \centering
  \includegraphics[width=\textwidth]{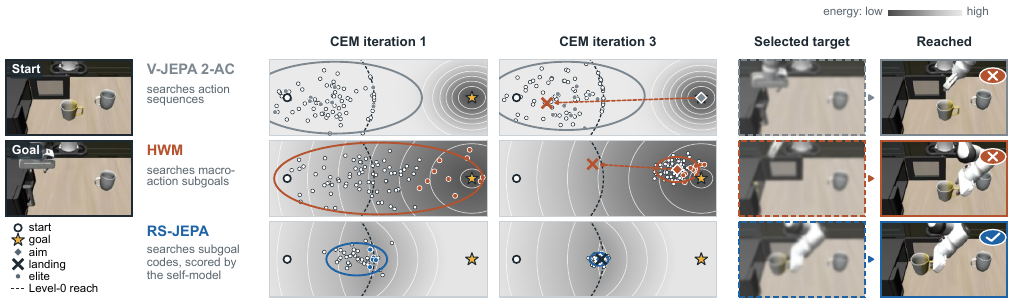}
  \caption{\textbf{Search comparison on one LIBERO-Long query.} The figure uses frames from one RS-JEPA episode and shows a shared two-dimensional view shaded by energy, with the reach of Level 0 dashed. Flat CEM on a distant goal samples action sequences whose energies barely differ, so elite selection carries little signal and its distribution stays broad, as \cref{eq:critical} of the main paper describes. HWM converges to a low-energy subgoal beyond reach and lands short of it. The prior of RS-JEPA starts the search near reachable codes, so three iterations with few samples suffice, and the selected subgoal matches the executed landing.}
  \label{fig:search}
\end{figure*}

\begin{table*}[tp]
  \begin{minipage}[t]{\columnwidth}
    \centering
    \caption{\textbf{Backbone analysis.} Success rate (\%) on LIBERO under the language protocol for two JEPA world models. On each backbone, a hierarchy improves on flat planning and the self-model improves further.}
    \label{tab:backbone}
    \small
    \setlength{\tabcolsep}{3pt}
    \begin{tabular}{lcccc}
      \toprule
      Method & \shortstack{LIBERO-\\Spatial $\uparrow$} & \shortstack{LIBERO-\\Object $\uparrow$} & \shortstack{LIBERO-\\Goal $\uparrow$} & \shortstack{LIBERO-\\Long $\uparrow$} \\
      \midrule
      \multicolumn{5}{l}{\emph{V-JEPA 2-AC backbone}} \\
      V-JEPA 2-AC~\cite{assran2025vjepa2} & 48.4 & 59.9 & 42.7 & 11.6 \\
      HWM~\cite{zhang2026hwm} & \underline{54.3} & \underline{60.6} & \underline{49.0} & \underline{20.3} \\
      RS-JEPA (Ours) & \textbf{63.9} & \textbf{62.3} & \textbf{61.0} & \textbf{50.7} \\
      \midrule
      \multicolumn{5}{l}{\emph{LeWM backbone}} \\
      LeWM~\cite{maes2026lewm} & 20.5 & 30.2 & 17.3 & 2.3 \\
      Hi-LeWM~\cite{caselli2026hilewm} & \underline{26.0} & \underline{32.1} & \underline{21.8} & \underline{10.8} \\
      RS-JEPA (Ours) & \textbf{34.8} & \textbf{38.9} & \textbf{29.4} & \textbf{24.8} \\
      \bottomrule
    \end{tabular}
  \end{minipage}%
  \hfill
  \begin{minipage}[t]{\columnwidth}
    \centering
    \caption{\textbf{Long-horizon split.} Success rate (\%) on LIBERO-Long, with the ten tasks grouped by demonstration length into the 3 shortest, the 3 middle and the 4 longest. The margin over the strongest baseline grows with task length.}
    \label{tab:split}
    \small
    \begin{tabular}{lccc}
      \toprule
      & \multicolumn{3}{c}{LIBERO-Long (\%), by length} \\
      \cmidrule(lr){2-4}
      Method & Short $\uparrow$ & Medium $\uparrow$ & Long $\uparrow$ \\
      \midrule
      V-JEPA 2-AC & 16.8 & 11.3 & 7.9 \\
      HWM & 28.1 & 22.5 & 12.8 \\
      H-JEPA & 32.3 & 25.4 & 16.7 \\
      Metro-WM & \underline{44.4} & \underline{41.2} & \underline{28.8} \\
      SAGE & 37.8 & 34.9 & 22.5 \\
      RS-JEPA (Ours) & \textbf{55.3} & \textbf{52.4} & \textbf{46.0} \\
      \bottomrule
    \end{tabular}
  \end{minipage}
\end{table*}

\begin{table*}[tp]
  \centering
  \caption{\textbf{Implementation ablations} on LIBERO-Long and LIBERO-Goal, with the aim--arrive gap on LIBERO-Long. The uncertainty penalty and weight sharing each add a smaller gain than the components of the main ablation.}
  \label{tab:ablimpl}
  \small
  \setlength{\tabcolsep}{4pt}
  \begin{tabular}{lccc}
    \toprule
    Variant & \shortstack{LIBERO-\\Long $\uparrow$} & \shortstack{LIBERO-\\Goal $\uparrow$} & Gap $\downarrow$ \\
    \midrule
    \textbf{Full model} & \textbf{50.7} & \textbf{61.0} & \textbf{0.20} \\
    \midrule
    w/o uncertainty penalty & 49.0 & 59.9 & 0.26 \\
    Shared $\to$ per-level weights & \underline{49.4} & \underline{60.2} & \underline{0.23} \\
    \bottomrule
  \end{tabular}
\end{table*}

\begin{table*}[tp]
  \centering
  \caption{\textbf{Additional benchmark results.} Success rate (\%) under offline language goals and full-interaction image goals. \emph{Offline}: language goals with no interaction, so no Dyna fine-tuning for the baselines and no self-improvement rounds for RS-JEPA. \emph{Goal image}: every suite with goal images and the full interaction budget. Metro-WM leads on LIBERO-Object and RS-JEPA leads on the other four suites. \emph{Reference}: FF-JEPA needs no goal and is not ranked.}
  \label{tab:offline}
  \small
  \begin{tabular}{lccccc}
    \toprule
    Method & LIBERO-Spatial $\uparrow$ & LIBERO-Object $\uparrow$ & LIBERO-Goal $\uparrow$ & LIBERO-Long $\uparrow$ & RoboCasa $\uparrow$ \\
    \midrule
    \multicolumn{6}{l}{\emph{Offline, language goal}} \\
    V-JEPA 2-AC & 47.4 & 58.4 & 42.1 & 11.0 & 17.9 \\
    HWM & \underline{52.7} & 57.6 & 46.4 & 16.8 & 19.9 \\
    H-JEPA & 49.5 & 54.3 & \textbf{46.7} & 20.8 & \underline{22.9} \\
    SAGE & 46.8 & 52.4 & 44.0 & \underline{26.3} & 21.8 \\
    Metro-WM & \textbf{54.3} & \underline{58.7} & \underline{46.6} & \textbf{33.6} & \textbf{26.2} \\
    RP1 & 45.1 & \textbf{58.9} & 41.4 & 10.4 & 18.0 \\
    RS-JEPA, round 0 & 47.8 & 57.6 & 40.6 & 11.5 & 19.1 \\
    \midrule
    \multicolumn{6}{l}{\emph{Goal image}} \\
    V-JEPA 2-AC & 49.9 & 58.2 & 44.2 & 13.1 & 19.8 \\
    HWM & 55.3 & 61.4 & 47.8 & 21.2 & 21.4 \\
    H-JEPA & 52.7 & 58.9 & 49.6 & 24.5 & \underline{27.2} \\
    SAGE & 52.4 & 57.5 & 46.2 & 32.6 & 26.4 \\
    Metro-WM & \underline{57.6} & \textbf{64.4} & \underline{50.4} & \underline{36.5} & 27.0 \\
    RP1 & 45.3 & 62.5 & 41.8 & 14.3 & 18.4 \\
    RS-JEPA (Ours) & \textbf{64.0} & \underline{63.6} & \textbf{61.5} & \textbf{51.9} & \textbf{31.1} \\
    \midrule
    \multicolumn{6}{l}{\emph{Reference}} \\
    FF-JEPA~\cite{masip2026ffjepa} & 48.8 & 59.9 & 39.1 & 15.8 & 21.8 \\
    \bottomrule
  \end{tabular}
\end{table*}

\begin{figure*}[tp]
  \centering
  \includegraphics[width=0.92\textwidth]{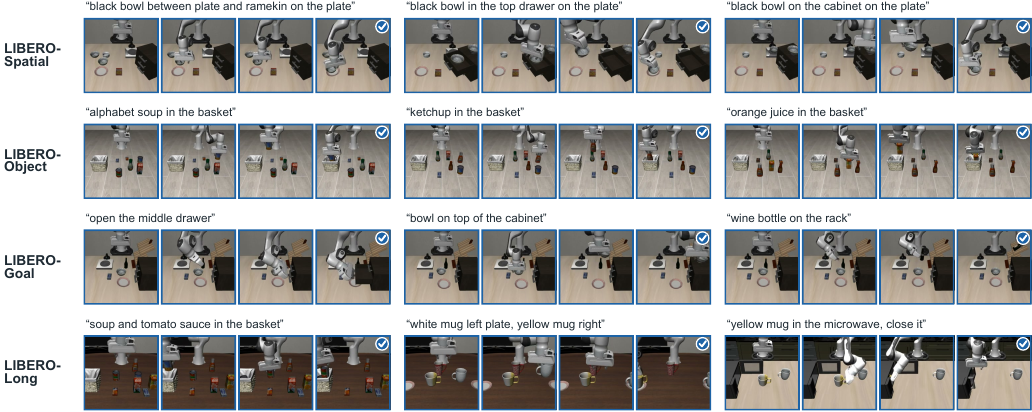}
  \caption{\textbf{Qualitative results on LIBERO.} One RS-JEPA episode for each of three tasks per suite, with four frames from the start to success; quotes abbreviate the task instructions. On the two-stage LIBERO-Long tasks, the middle frames fall in the first and the second stage.}
  \label{fig:qual}
\end{figure*}

\begin{figure*}[tp]
  \centering
  \includegraphics[width=0.92\textwidth]{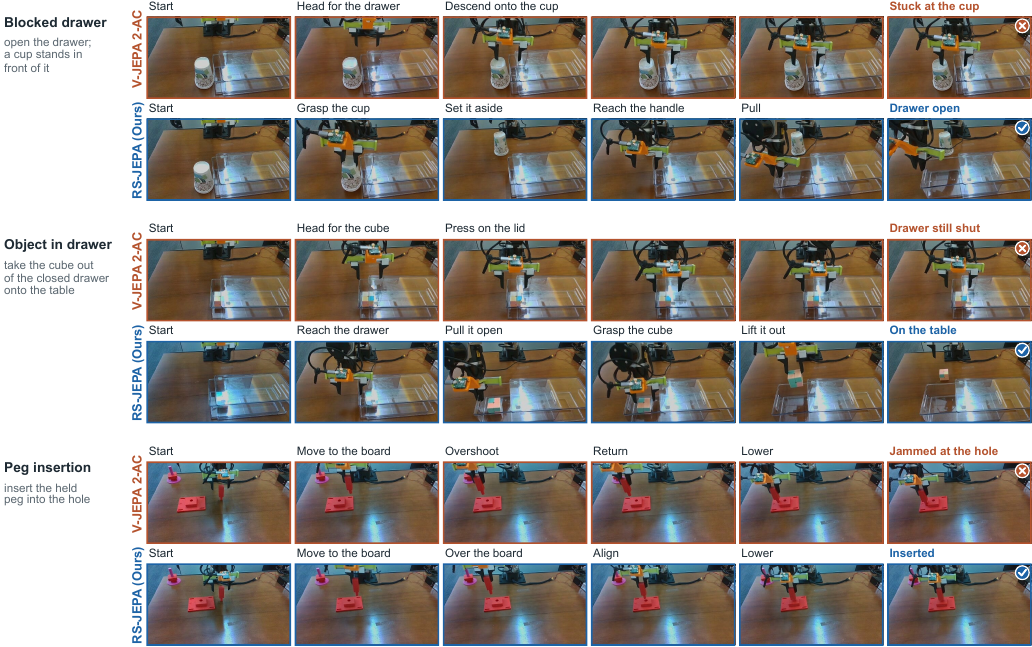}
  \caption{\textbf{Real-robot tasks that need a detour.} V-JEPA 2-AC (top) and RS-JEPA (bottom) from the same start on the B601-RS arm, with each frame labelled by the phase it shows. Blocked drawer: RS-JEPA first sets the cup aside, while V-JEPA 2-AC heads for the drawer and stalls on the cup. Object in drawer: RS-JEPA opens the drawer before it grasps the cube, while V-JEPA 2-AC reaches for the cube through the closed lid. Peg insertion: V-JEPA 2-AC overshoots the board and jams the peg beside the hole.}
  \label{fig:qualdetour}
\end{figure*}

\begin{figure*}[tp]
  \centering
  \includegraphics[width=0.92\textwidth]{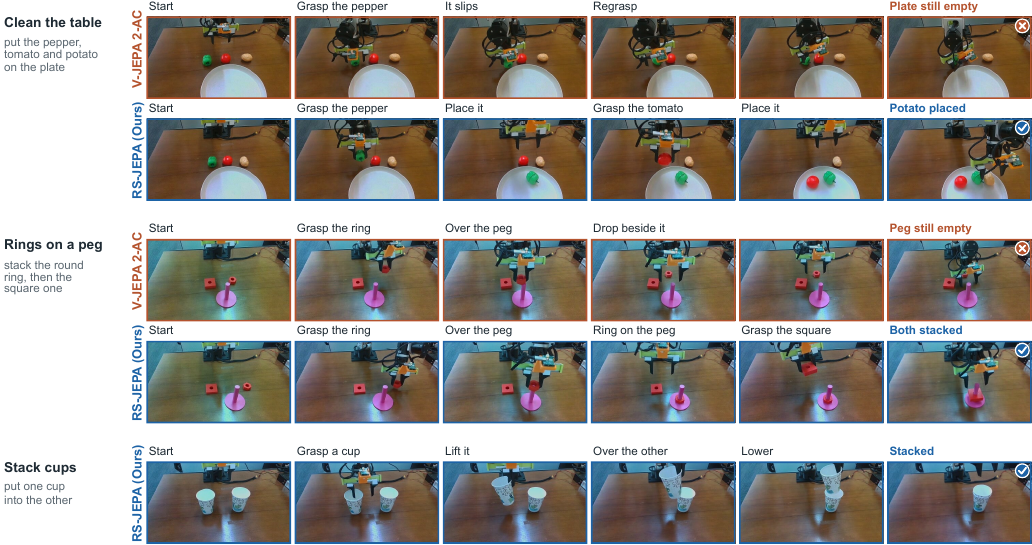}
  \caption{\textbf{Long-horizon real-robot tasks.} In paired rows, V-JEPA 2-AC (top) and RS-JEPA (bottom) start from the same state. Clean the table: RS-JEPA places the three objects in turn, while V-JEPA 2-AC loses its grasp on the first. Rings on a peg: RS-JEPA stacks both rings in order, while V-JEPA 2-AC drops the first ring beside the peg. Stack cups is the short-horizon control, shown for RS-JEPA.}
  \label{fig:quallong}
\end{figure*}

\fi

\end{document}